\documentclass[10pt, journal,final,twocolumn]{IEEEtran}

\newcommand{\1} {{\bf 1}} 
\newcommand{\I} {{\bf I}} 
\newcommand{\G} {{\bf G}} 

\usepackage{lineno}
\usepackage{graphics}
\usepackage{epsfig}
\usepackage{graphicx} 
\usepackage{amssymb}
\usepackage{amsmath}
\usepackage{algorithmic}
\usepackage[ruled]{algorithm2e}

\begin{document}

\markboth{IEEE TRANSACTIONS ON IMAGE PROCESSING, DOI (identifier) 10.1109/TIP.2012.2211373}{}

\title{Integrating Graph Partitioning and Matching for Trajectory Analysis in Video Surveillance}
%
%
\author{Liang Lin, Yongyi Lu, Yan Pan, Xiaowu Chen\thanks{ 
Copyright (c) 2012 IEEE. Personal use of this material is
permitted. However, permission to use this material
for any other purposes must be obtained from the IEEE by
sending an email to pubs-permissions@ieee.org. This work was supported by the National Natural Science Foundation of China (Grant No.61173082), the Hi-Tech Research and Development Program of China (National 863 Program, Grant No.2012AA011504), and the Guangdong Natural Science Foundation (Grant No.S2011010001378). This work was also supported in part by the open funding project of State Key Laboratory of Virtual Reality Technology and Systems, Beihang University (Grant No. BUAA-VR-12KF-06).

L. Lin, Yongyi Lu, and Yan Pan are with the Sun Yat-Sen University, Guangzhou
510275, China. (e-mail: linliang@ieee.org).

Xiaowu Chen is with the School of Computer Science and Engineering, Beihang University, Beijing 100191, China. (e-mail: chen@buaa.edu.cn).

}}

%
%
%


\maketitle

\begin{abstract}
In order to track the moving objects in 
long range against occlusion, interruption, and background
clutter, this paper proposes a unified approach for global trajectory analysis. Instead of the traditional frame-by-frame tracking, our method recovers target trajectories based on a short sequence of video frames, e.g. $15$ frames. We initially calculate a foreground map at each frame, as obtained from a state-of-the-art background model. An attribute graph is then extracted from the foreground map, where the graph vertices are image primitives represented by the composite features. With this graph representation, we pose trajectory analysis as a joint task of spatial graph partitioning and temporal graph matching. The task can be formulated by maximizing a posteriori under the Bayesian framework, in which we integrate the spatio-temporal contexts and the appearance models. The probabilistic inference is achieved by a data-driven Markov Chain Monte Carlo (MCMC) algorithm. Given a peroid of observed frames, the algorithm simulates a ergodic and aperiodic Markov Chain, and it visits a sequence of solution states in the joint space of spatial graph partitioning and temporal graph matching. In the experiments, our method is tested on several challenging videos from the public datasets of visual surveillance, and it outperforms the state-of-the-art methods.
\end{abstract}


\IEEEpeerreviewmaketitle

\section{Introduction}
\label{sect:introduction}

\begin{figure*}
\centering
\includegraphics[width=0.8\linewidth]{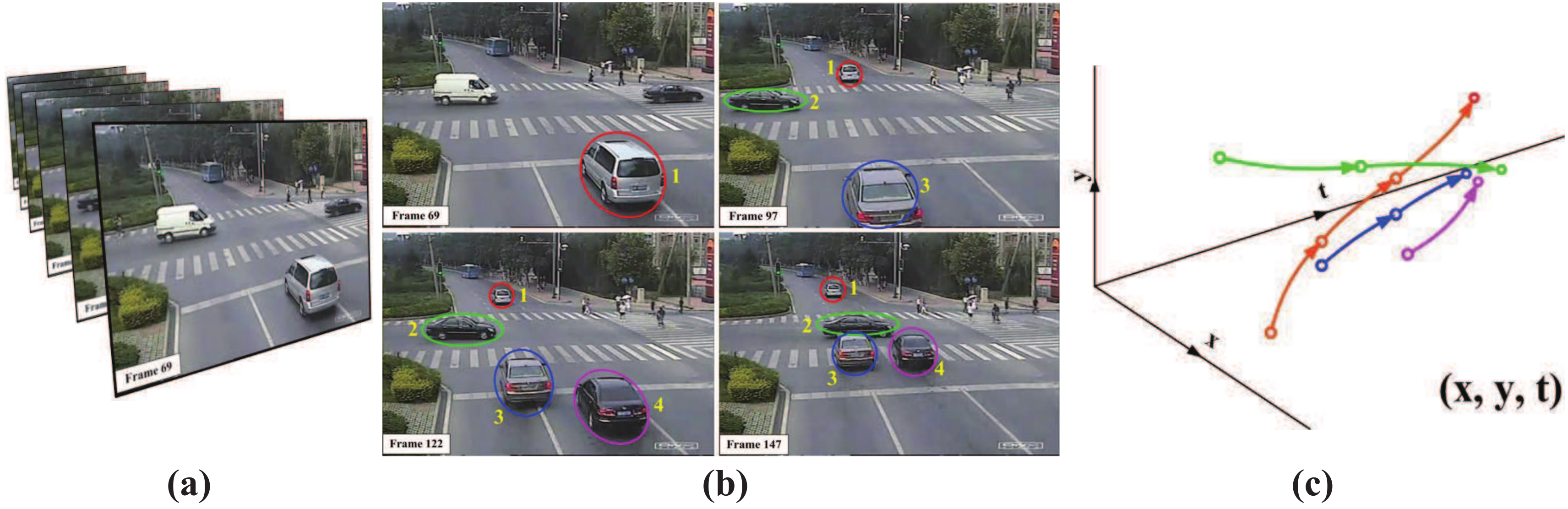}
\caption{Illustration of the trajectory analysis. (a) shows a batch of successive video frames as the input of our method. (b) shows a few results of multiple target tracking, where the numbers around the tracking ellipses imply the identities of targets. (c) visualizes the global trajectories of the video in a 3D perspective.} \label{fig:front_fig}
\end{figure*}

Video object tracking is a fundamental problem in the academic research of image/video processing and computer vision, involving two key issues: (i) extracting objects of interest from backgrounds and (ii) establishing correspondences of objects over video frames. Trajectory parsing and analysis for multiple targets is a further task upon target tracking, and plays a critical role in the recently-arising intelligence applications, such as robotics~\cite{SMC-Robot} and video surveillance systems~\cite{SurveillanceSystem,TIP-Surveillance,TIP-SeqParticle}. It is also an important support for higher level video retrieval and event analysis~\cite{LinEvent,LHIDataset}. The object of this work is to study a unified approach for trajectory analysis under the Bayesian framework. As Fig.~\ref{fig:front_fig} illustrates, the input of our algorithm is a short sequence of observed frames rather than a single frame, in which we localize the multiple moving targets and track them with their identities preserved; the global trajectories of targets for the whole video can be parsed through the inference. 


\subsection{Related Work}

In the literature, video object tracking has been intensively studied and many effective methods have been proposed. For single-target tracking, various object appearance models and motion models are well exploited to estimate target state (location, velocity, etc.)~\cite{TrackingSurvey,SMC-Kalman,LinTrackingAR,LearnAssociate2009,TIP-SeqParticle}. Recently, a class of techniques called ``tracking by detection'' has been shown to provide promising results~\cite{MILTrack,AdaptiveTracking,SMC-DecisionTrack,SMC-PartTracking,TIP-AdaptiveTracking}. For multi-object tracking (i.e. trajectory analysis), which our method addresses, we shall identify multiple moving targets by associating correspondences between observations and objects as well as estimating the state of each target~\cite{TrackingSTContext,TrajectoryParsing}.

In general, we roughly categorize the work of trajectory analysis into two types: sequential inference based, and deferred inference based, in terms of the number of input frames for inference. 

(I) Sequential inference based methods use the information of the currently observed frame to predict the states of moving targets and assign their target identities. The classical examples are particle filtering~\cite{MCMCParticleFiltering,TIP-SeqParticle,SMC-Kalman} and optical flow~\cite{OpticalFlow}. Recently, Avidan~\cite{EnsembleTrack} proposed a learning-based tracker using the online Adaboost algorithm, which maintains a discriminative detector to track targets in the current frame. Babenko et al.~\cite{MILTrack} significantly improved the tracking performance using Multiple Instance Learning (MIL). Despite great success, these approaches may yield identity lossing (or switching) and trajectory fragmentation in terms of mutual-interaction, occlusion and spurious motion, because they make online decisions while discarding global information.

\begin{figure}[!ht]
\begin{center}
   \includegraphics[width=0.6\linewidth]{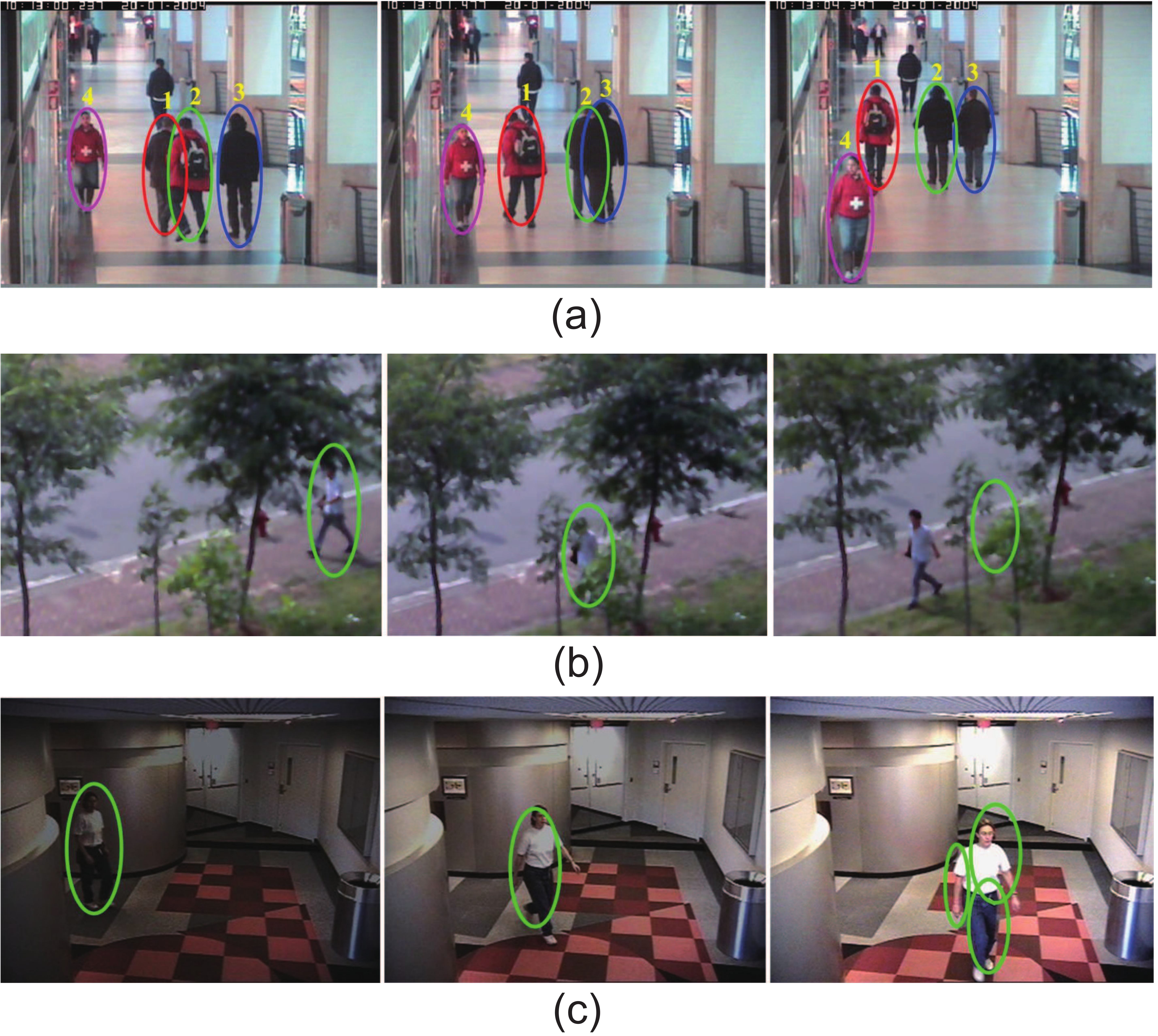}
\end{center}
\caption{A few typical challenges in trajectory analysis. (a) Due to the mutual interaction in the crowd scene, the track IDs of targets are switched. (b) The tracker is distracted by the background clutter. (c) The tracked target is split into several ones, due to illumination and object scale changes.} \label{fig:challenge}
\end{figure}

(II) Deferred inference based methods, also referred as global data association based tracking, are to identify each observation with either a track ID or a false alarm in a short period of time, e.g. $15$ frames. The observations, namely, moving blobs, can be obtained by using methods such as background subtraction. The first attempts on data association optimization are Multiple Hypothesis Tracker~\cite{MHT,SMC-Kalman,SMC-PartTracking}, and Joint Probabilistic Data Association Filters~\cite{JPDAF}, which search the hypothesis (the associations of observations and targets) by assuming one-to-one mapping, i.e. one observation to one target. Once this assumption is relaxed, e.g. a target consisting of a set of observations, the search space of optimization grows exponentially with the number of frames and targets. To overcome this problem, many deterministic optimal algorithms have been employed, such as Extended Dynamic Programming~\cite{TrackDP,SMC-DynamicProg,TIP-SeqParticle}, Quadratic Boolean Programming~\cite{TrackQBP}, and Hierarchical Hungarian algorithm~\cite{TrackHieHung}. However, it is still impractical to apply these methods for intelligence surveillance systems, due to the following aspects~\cite{TIP-Surveillance,TIP-MultipleTrack,SurveillanceSystem}. First, some approaches of trajectory analysis need good initializations, e.g., manually annotating targets or assuming no conglutination at the beginning frame. Second, due to the ambiguity caused by the similar appearances of coupled targets, it is difficult to stably maintain the correct identities of targets with long term tracking. In the example in Fig.~\ref{fig:challenge} (a), the track IDs of targets are switched in the crowd scene~\cite{SMC-Kalman}. Third, the affinity model of a moving target, i.e. object representation, is not discriminative with respect to complex surrounding clutter, illumination and object scale changes, which often leads to false tracking or the splitting of one target into several pieces~\cite{SMC-PartTracking,TIP-MultipleTrack}, as the examples shown in Fig.~\ref{fig:challenge} (b) and (c).

\subsection{Method Overview}

According to the literature review, the proposed approach belongs to deferred inference based methods. The goal of our approach is to parse trajectories of moving targets under the Bayesian framework, in which searching for the optimal trajectory solution is formulated as a problem of maximizing a posterior probability (MAP). We briefly introduce our method in the following three aspects: a composite feature for matching affinity of moving targets, a spatio-temporal graph for representing the task of trajectory analysis, and an iterative stochastic algorithm for global inference. 

(I) In surveillance videos, particularly for some outdoor scenes, it is a critical issue to robustly recover correspondences over frames against illumination changes, drastic motion, etc. A consensus from a recent image feature research~\cite{TuytelaarsFeatureSurvey} is that a good image feature for tracking demands two properties: (i) the discrimination, i.e. distinctive matching over frames, and (ii) the robustness, i.e. geometric-invariance, and tolerance of non-rigid motion, etc. In fact, these two properties sometimes conflict with each other. For example, one may increase the region size (scale) of a local feature and/or the dimensionality of the descriptor, but a larger feature is usually less robust in tracking with photometric and geometric changes. In this paper, we propose a composite image feature to represent moving targets. We employ two types of well-known image features, SURF~\cite{SURF} and MSER~\cite{MSER}, in the composite features. Each composite feature is composed of a feature region generated by MSER detector within a set of SURF feature points. This scheme is similar with the Bundled Feature~\cite{BundledFeature} proposed by Sun et al.~\cite{BundledFeature} for web image search, but we define a different matching metric to adapt object tracking.

(II) Given the extracted composite features from the observed frames, we can build up a spatial graph and a temporal graph to pose the problem of trajectory analysis as a joint task of spatial graph partitioning and temporal graph matching. In the spatial graph, each graph vertex is a detected composite feature and each graph edge is defined by the appearance and motion consistency of the two adjacent vertices. In the temporal graph, each graph vertex implies one underlying target consisting of a connected cluster of composite features, and the graph edges denote the matching correspondences between targets in consecutive frames. With these graph representations, the task of graph partitioning corresponds with extracting and segmenting targets from background; the graph matching task is equivalent to establishing the correspondences of targets over frames. We can further formulate these tasks by maximizing posterior probability under the Bayesian framework. In addition, two types of scene contexts are integrated as the informative prior, including: (i) target size prediction using scene geometric information, inspired by the previous work ~\cite{SurveillanceSystem,TrackingSTContext}, and (ii) target motion prior model by the path statistics. These types of prior knowledge are very informative to make the model robust and efficient. For example, with two people walking close together with similar appearances, our model tends to segment them into two individual targets according to the prior term of target size prediction.

(III) It is a non-trivial optimization procedure to search for the maximum of the posterior probability with our formulation. There are many ambiguities caused by conglutinations, occlusions, and similar appearances of targets and background clutters in some crowded surveillance scenes. The searching order or rule for an optimal solution is thus quite difficult to design. In the perspective of energy minimization, there exists quite a few local minimums, e.g., track ID switching, in the search for energy minimums. Therefore, unlike the deterministic or heuristic searching in the previous work of trajectory inference~\cite{TrackDP,TrackingSurvey}, we design a stochastic sampling algorithm using the Markov Chain Monte Carlo (MCMC) mechanism~\cite{Metropolis} to explore the solution space. In literature, some work~\cite{MCMCDataAssociation} shows great results on solving spatio-temporal data association by an MCMC-based algorithm. In our method, we adopt an MCMC-based cluster sampling method, namely Swendsen-Wang Cut~\cite{SWCBarbu}, for optimal solution exploration. The algorithm iterates between two types of MCMC dynamics for the spatial graph partitioning and temporal graph matching respectively.


Compared with some recently proposed approaches~\cite{TrajectoryParsing,MCMCDataAssociation} which also adopt stochastic inference for trajectory analysis, the major advantages of the proposed method are as follows. (1) We adopt two types of MCMC dynamics to iteratively solve the video object segmentation and tracking, which are mutually conditional and closely coupled. This algorithm is able to explore the global optimal solution and eliminate the need for good initializations. (2)  The proposed composite feature provides a flexible and robust representation against scene clutters and object geometric deformations in tracking. (3) We apply our method to various challenging surveillance videos from several public datasets and show that it outperforms other approaches.


This paper is organized as follows. We first introduce the
problem representation and formulation in Section~\ref{sect:representation} and Section~\ref{sect:formuation}. Then Section~\ref{sect:inference} presents the algorithm for trajectory inference, and Section~\ref{sect:implementation} describes the implementation details and the system flow. A set of experiments and comparisons are proposed in Section~\ref{sect:experiments}, and the paper is concluded with discussions in Section~\ref{sect:discussion}.





\section{Problem Representation}
\label{sect:representation}

Given an input video, we set the observed window spanned over $\tau$ frames for each computation of trajectory analysis. The observed window is moving with a step-size of $\eta$ frames. Using a state-of-the-art background modeling algorithm~\cite{BackgroundModel}, the image lattice $\Lambda_t, t= 1, \ldots, \tau$ of each frame is initially partitioned into foreground and background domains $\Lambda_t=\Lambda^B_t \cup \Lambda^F_t$. The trajectory analysis takes the foreground domain as the input, although the background subtraction is not perfect, i.e. occurring false alarm regions. We then propose a novel image feature, namely the composite feature, extracted from the foreground domain, based on which a spatial graph and a temporal graph are constructed. Each vertex in the spatial graph is a composite feature and each vertex in the temporal graph represents a segmented moving target. In the following, we start by introducing the composite features, then define the problem of trajectory analysis via graph representation, and present the probabilistic formulation.

\begin{figure}[t!]
\begin{center}
   \includegraphics[width=0.6\linewidth]{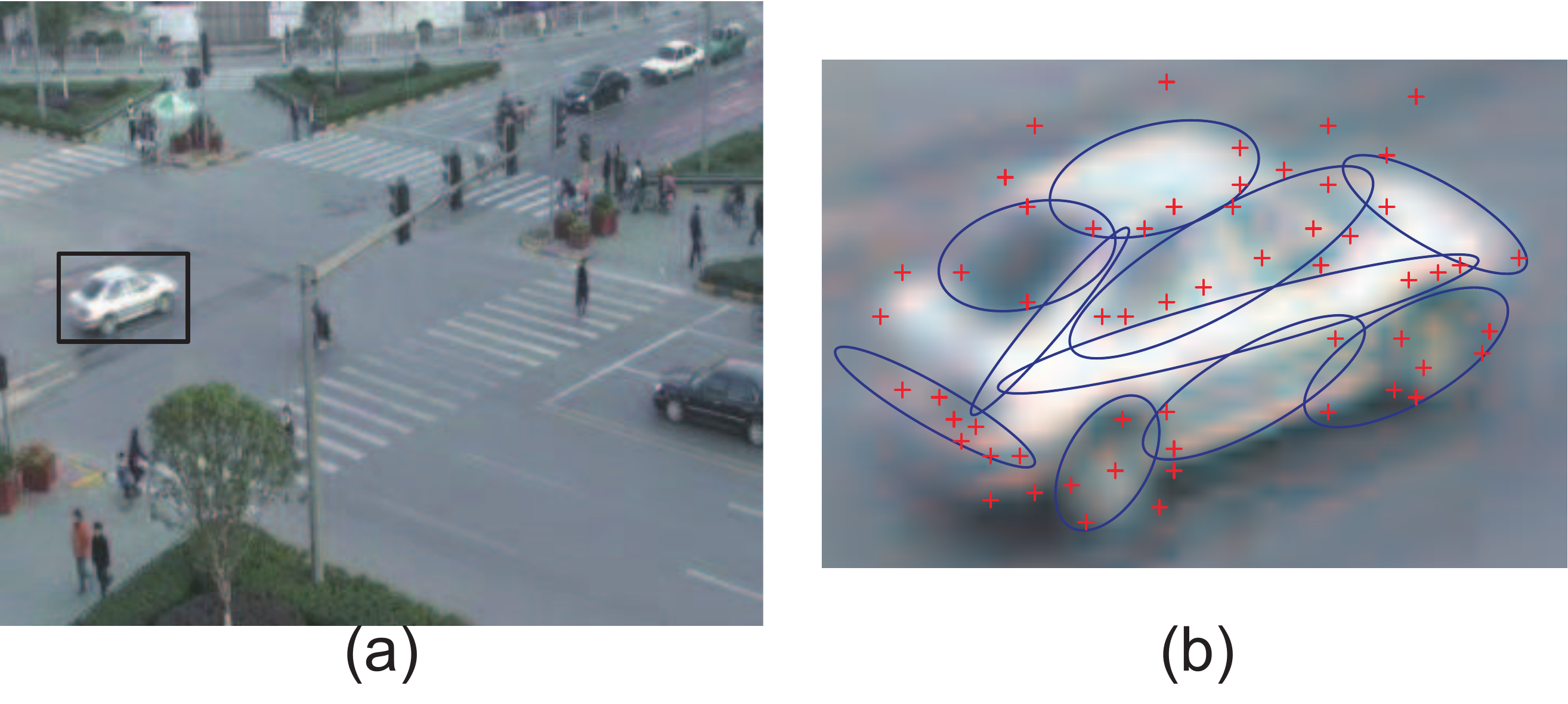}
\end{center}
\caption{The composite feature bundling SURF points and MSER regions. The moving target tracked by a black bounding box in (a) can be represented by the composite features in (b), where the blue ellipses indicate the MSER regions and the red crosses indicate the SURF points. Note we discard the MSER regions having heavy overlap or without SURF points included.} \label{fig:comp_feature}
\end{figure}

\subsection{Composite Features}
\label{sect:comp_features}

For representing moving objects, we propose a composite image feature that bundles a region with several key points for improving both discrimination and robustness. The proposed composite feature involves two popular features: the point feature SURF~\cite{SURF} and the region feature MSER~\cite{MSER}. The SURF keypoint exploits scale-space extrema by determination of Hessian matrix and employs integral image for rapid computation. The MSER feature is defined by an extremal property of its intensity function in the ellipse region and on its outer boundary. Both of these two features are robust against viewing angle, scale, and illumination changes. Some extracted SURF points and MSER regions are shown in Fig.~\ref{fig:comp_feature} (b).

Given a foreground image domain $\Lambda^F_t$, we first detect the point and region features, denoted by $ S = \{ s_i \}$ and $ R = \{ r_j \}$ respectively. We allow overlaps among the region features, and discard those with large size, i.e. those containing others or spanning half the size of the foreground domain. A composite feature $Z_j$ is then defined as 
\begin{equation}
Z_j \!\!=\!\! \{ r_j , S_j \!\!=\!\! \{ s_i : s_i \propto r_j, s_i \in S \} \}, r_j \in R, S_j \subset S,
\end{equation}
where $s_i \propto r_j$ indicates that the point feature $s_i$ exists inside the region feature $r_j$. The composite feature including no SURF points will be removed automatically. In practice, the number of SURF points in each composite feature is $5 \sim 10$. A moving target represented by the composite features is illustrated in Fig.~\ref{fig:comp_feature}.

The measuring energy $E(Z_a, Z_b)$ of two composite features $Z_a$ and $Z_b$ includes two terms: {\em independence similarity } and {\em configuration consistency}.
\begin{equation}\label{eqn:feature_measure}
E (Z_a, Z_b) = E_{I} + \lambda_g E_{G},
\end{equation}
where $\lambda_g$ is a weighted parameter for the two terms.

(I) The independence similarity $ E_{I} $ is based on the matching distance of two region features. The energy of this term is defined as,
\begin{equation}\label{eqn:ind_similarity}
E_{I} (Z_a, Z_b) =  \| h(r_a) - h(r_b) \| ^2 ,
\end{equation}
where $h(\cdot)$ is the descriptor for SURF feature. 

\begin{figure}[t!]
\begin{center}
   \includegraphics[width=0.4\linewidth]{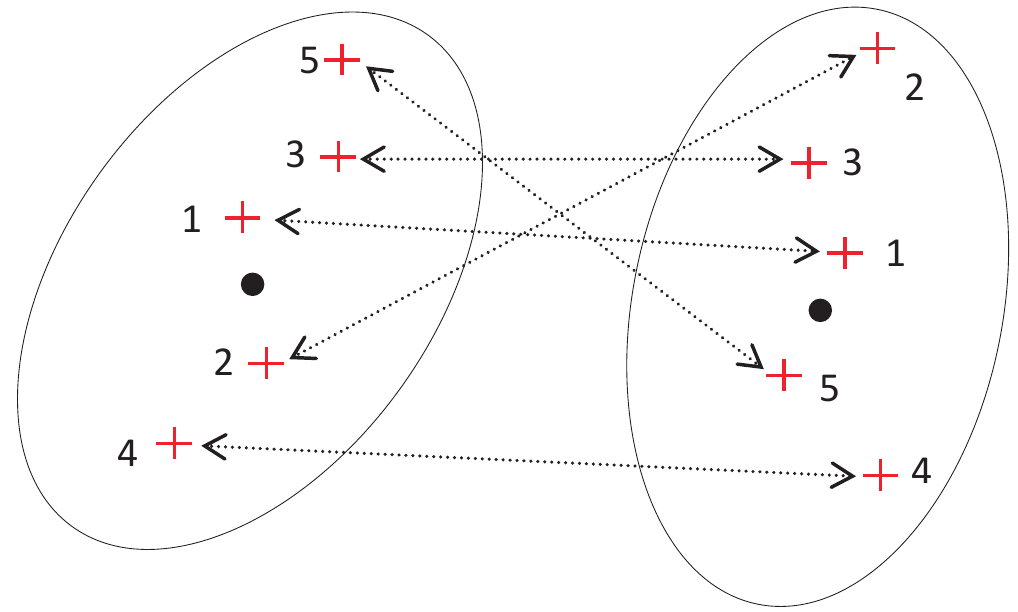}
\end{center}
\caption{An example of the measuring configuration consistency of two composite features. We denote the MSER region by the ellipse, the SURF points by the red cross, and the centroid of the feature by the black spot. For the left composite feature, its relative order for the configuration is: $\{ 1,  2,  3,  4,  5 \}$, and for the right one, its relative order is $\{ 1,  5,  3,  4,  2 \}$. Thus, the configuration consistency of these two composite features is: $ \frac {1  +  0 +  1 + 1 + 0} {5} =  0.6$.} \label{fig:match_comp_feature}
\end{figure}

(II) The configuration consistency $ E_{G} $ performs a weak
geometric verification between two composite features. Let $\{ s_i \leftrightarrow s_j, s_i \in S_a, s_j \in S_b \}$ denotes the set of matched feature pairs of two composite features $Z_a$ and $Z_b$. This set can be quickly calculated by matching SURF points in a greedy manner: searching the best match for each point in region of the corresponding composite feature. We define their configuration consistency based on the relative order with point matching. Given the centroid of region feature, the relative order of inside points can be determined according to their spatial distance to the centroid. As Fig.~\ref{fig:match_comp_feature} illustrates, we number the points in the left based on the spatial distance to the centroid, i.e. {1,2,3,4,5}; the numbers of points in the right is propagated from the left points based on the matching correspondence. And the consistency can be computed as,
\begin{equation}\label{eqn:conf_similarity}
E_{G} (Z_a, Z_b) = \frac {\sum_{s_i \leftrightarrow s_j} {\1} ( \mathcal{O} (s_i) =  \mathcal{O} (s_j) )} {|\{ s_i \leftrightarrow s_j \}|} ,
\end{equation}
where $ \mathcal{O}$ denotes the relative order of the points, ${\1} (\cdot)$ is the indicator function, and $|\{ s_i \leftrightarrow s_j \}|$ is the number of matched point pairs. The unmatched point pairs are not taken into account in the definition because the appearance dissimilarity has been penalized by the first term $E_{I}$ in Eqn.~\ref{eqn:ind_similarity}. Specifically, the cost by $E_{I}$ would be relatively large with respect to the $E_G$, if the numbers of points are discrepant (e.g., 5 v.s. 10). Moreover, to make this consistency penalty smooth and gentle, we can additionally apply the {\em sigmoid} function on the relative order computation.

We observe that, unlike a single type of features, a composite feature provides a flexible and stable representation that captures the distinctive image primitives as well as the geometric structure.

\subsection{Trajectory Analysis via Graph Representation}

Given the observed window, i.e. a period of frames $\I_{[0, \tau]}$, we extract the composite features $\{ Z_{t,i}, t = 0, \ldots, \tau \}$ on the foreground areas $\Lambda^F_{[0, \tau]}$. Then we obtain a set of spatial graphs $\G^S_{[0, \tau]}$, where each composite feature is treated as the graph vertex $ v_{t,i} = Z_{t,i}, t = 0, \ldots \tau $. 

The goal of trajectory analysis is to segment moving targets and recover their correspondences in each frame. With the graph representation, this problem is posed as a joint task of graph partitioning and matching.

\begin{figure}[t!]
\begin{center}
   \includegraphics[width=0.55\linewidth]{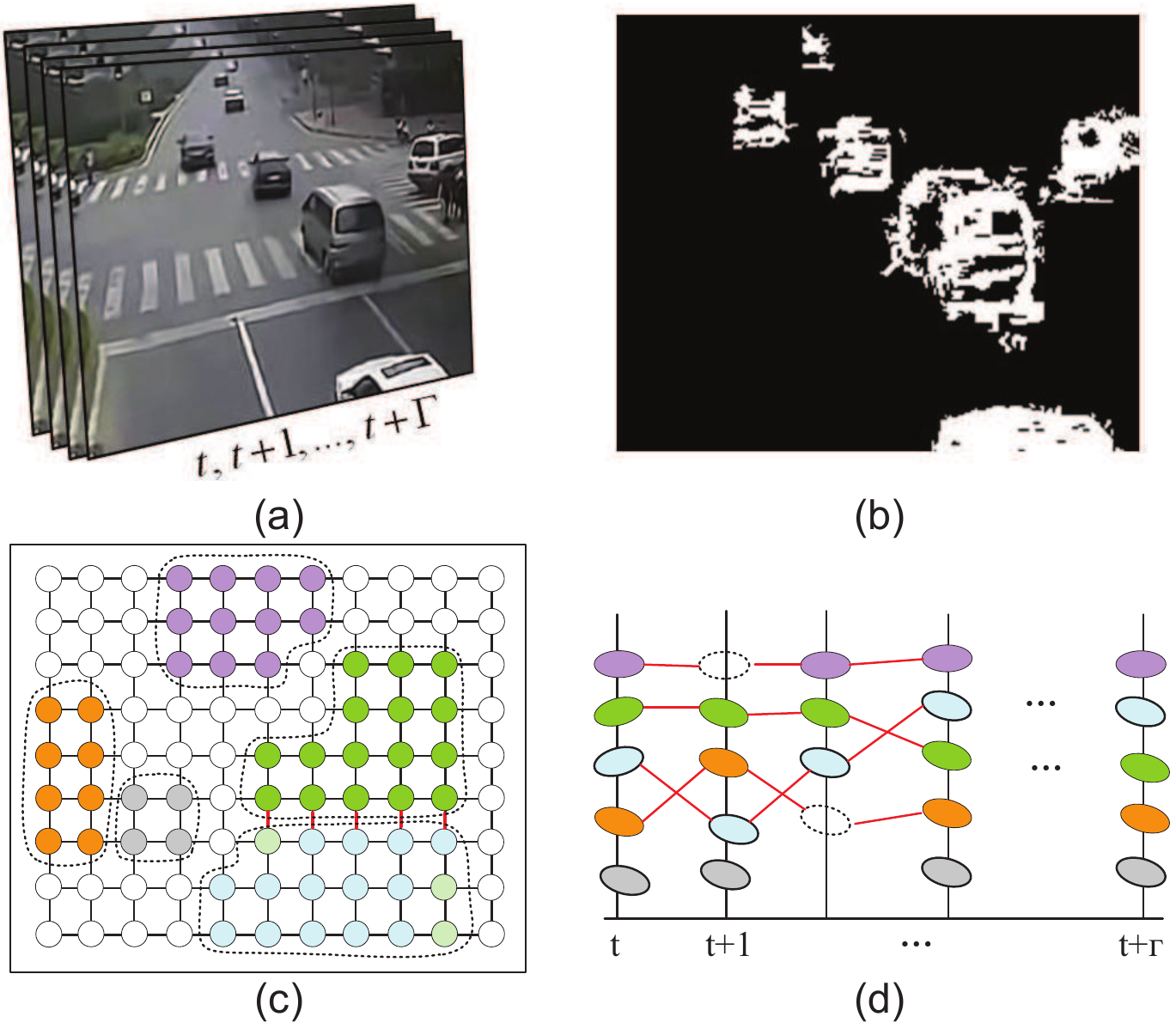}
\end{center}
\caption{The graph representations for trajectory analysis. (a) shows the input video sequence $I_{[t,t+\tau]}$. (b) shows the foreground mask for frame $I_t$. (c) illustrates the spatial attribute graph of the currently observed frame $I_t$, where each graph vertex denotes a composite feature of the foreground domain and has four bonds connecting to neighboring vertices. The graph edges imply the motion and appearance consistency between two adjacent vertices. The edges between the foreground and background domains are turned off automatically. (d) illustrates the temporal attribute graph with the vertices being the connected clusters of spatial graph vertices. Each temporal graph vertex indicates an underlying target. The edges in the temporal graph represent the matching correspondences over frames. Note that the vertices in the bottom row in (d) indicate the unmatched regions and have no temporal connections.} \label{fig:stgraph}
\end{figure}

\textbf{I. Spatial graph partitioning} is to segment targets over a
time span $\tau$. As illustrated in Fig.~\ref{fig:stgraph} (b), we 
represent the partition of the observed frames as $\Pi_{[0,\tau]}$,
\begin{eqnarray}\label{eq:partition}
\Pi_{[0,\tau]}=\{\pi_t; t=0,1,2,\ldots,\tau\} \nonumber \\
\pi_t=\{U_{t,i};i=0,1,2,\ldots,K_t\},
\end{eqnarray}
where $K_t$ is the target number at time $t$, and $U_{t,0}$ indicates the
false alarm regions, i.e. not target regions but proposed as the foreground. Each moving target $U_{t,i}$ at time $t$ is described
by a bounding box,
\begin{equation}\label{eq:target_represent}
U_{t,i}=\{x_{t,i},y_{t,i},w_{t,i},h_{t,i}\}, i=1,2,\ldots,K_t
\end{equation}
where $(x_{t,i},y_{t,i})$ denotes the target center and $(w_{t,i},h_{t,i})$ denotes the width and height. The initial foreground domain $\Lambda^F_t$ consists of the target image domains $\Lambda^F_{t,i}$ and false alarm domains $\Lambda^F_{t,0}$, 
\begin{equation}
\Lambda^F_{t} = \bigcup_{i=1}^{K_t} \Lambda^F_{t,i} \bigcup \Lambda^F_{t,0}.
\end{equation}

We solve the foreground partitioning $\Pi_t$ with a spatial graph representation (as shown in Fig.~\ref{fig:stgraph} (c)), defined over the foreground image lattice with nearest $4$ neighbor connections, $G_t^S=(V_t^S, E_t^S)$, where $V_t^S$ is the set of graph vertices and $E_t^S$ is the set of link edges connecting neighboring graph vertices. Each spatial graph vertex $v_{t,i}^S = (Z_{t,i}, l_{t,i}) \in V_t^S$ includes one composite feature $Z_{t,i}$ and the corresponding label $l_{t,i}=[0,K_t]$, indicating the vertex belongs to certain target or false alarm. Therefore, each target $U_{t,i}$ at time $t$ corresponds to a set of connected graph vertices $V^S_{t,i} \subset V^S_t$. We solve the task of graph partitioning by turning off edges, i.e., generating disjoint subgraphs, which will be introduced in Section~\ref{sect:spatial_sampling}.


%

\textbf{II. Temporal graph matching} is recovering the
correspondences of targets over time span $\tau$. We represent a set of matching matrices by $\Phi_{[0,\tau]}$,
\begin{eqnarray}
\Phi_{[0,\tau]} &=& \{\phi_t;t=1,2,\ldots,\tau-1\}
\\\nonumber \phi_t(U_{t,i}) &=& U_{t+1,i} \bigcup \{\emptyset\},
\end{eqnarray}
where each matrix $\phi_t$ describes a mapping relation from the $t$-th frame
to the $(t+1)$-th frame. A target matching to $\emptyset$ indicates that it is occluded or moving out at the current frame (i.e. being ``killed''), while a target with no matches in previous frames indicates that it is newly appearing (i.e. being ``born'').

As illustrated in Fig.~\ref{fig:stgraph} (d), a temporal graph
$G^T=(V^T,E^T)$ is defined for moving targets. Each temporal graph vertex
$v_{t,i}^T = (U_{t,i}, l_{t,i}) \in V^T$ includes a moving target $U_{t,i}$
and its matching label $l_{t,i}$ at time $t$. Each edge indicates the
matching relation of two vertices between adjacent frames, as $e_{t,i}
=\{<v_a, v_b>:  \; v_a, v_b \in V^T, \; <v_a, v_b> \in E_t^S \}$. Since we have performed partitioning on the spatial graph, we can reasonably assume one-to-one mapping between temporal nodes. Note that unmatched nodes are allowed to stand alone, caused by false alarm regions from the background subtraction. In Fig.~\ref{fig:stgraph}(d), the blobs with different colors represent the temporal graph nodes and the dotted ones indicate the unmatched regions.

Therefore, for the problem of trajectory completion, we define the following solution representation $W$ from the observed $I_{[0,\tau]}$ as
\begin{eqnarray}\label{eq:configuration}
W_{[0,\tau]}=\{K_{[0,\tau]},\Pi_{[0,\tau]},\Phi_{[0,\tau]}\},
\end{eqnarray}
where $K_{[0,\tau]}$ denotes the foreground target number in time span
$\tau$, $\Pi_{[0,\tau]}$ denotes the partition result for each frame, and $\Phi_{[0,\tau]}$ denotes the
matching correspondences of moving targets between adjacent frames in the
form of matrix mapping from one target to another.

Equivalently, the solution configuration of trajectory completion can
also be represented by $N$ motion trajectories, also called ``cables''  in~\cite{WangTrajectory},
\begin{eqnarray}
W_{[0,\tau]}=\{N,C_i;i=0,1,\ldots N\},
\end{eqnarray}
where $C_0$ represents the false alarm regions, and other cable
represents the trajectory of a foreground moving target, respectively. This representation
makes it simple to define the motion models.
\begin{eqnarray}
\!\!\!\!\!\!C_i\!\!\! &=& \!\!\!(t_{i,b},\!t_{i,d},\! \{U_{t,i}; t\in [
t_{i,b},t_{i,d}]\});  i \!=\! 1,\ldots,N, \\
\!\!\!\!\!\!C_0\!\!\! &=& \!\{U_{t,0};t=0,1,\ldots,\tau\},
\end{eqnarray}
where $t_{i,b}$ and $t_{i,d}$ denotes the birth time and death time of the trajectory $C_i$, respectively.

Therefore, in the probabilistic formulation in Section~\ref{sect:formuation}, we shall be able to switch between the two notations above.


\section{Probabilistic Formulation}
\label{sect:formuation}

Based on the definition of solution $W$, we can formulate the inference problem in a Bayesian framework, and the optimal solution $W^{*}$ can be solved by maximizing a posterior probability,
\begin{eqnarray}\label{eq:posterior}
W_{[0,\tau]}^* \!\! &=& \!\! \arg \max_W p(W_{[0,\tau]}|I_{[0,\tau]}) \\
&=&\!\!\!\! \arg \max_W p(I_{[0,\tau]}|W_{[0,\tau]};\beta)
p(W_{[0,\tau]}| \theta)\nonumber ,
\end{eqnarray}
where $\beta$ and $\theta$ are the parameters for the likelihood and prior models respectively.

\subsection{Prior model}\label{subsect:prior}

We define prior model $p(W_{[0,\tau]}| \theta)$ on scene contexts, which  provide informative guidance for graph partitioning and matching, as
\begin{equation}
p(W_{[0,\tau]}|\theta)\!\!\! = \!\!\!p(\Pi_{[0,\tau]}) \cdot p(\Phi_{[0,\tau]}).
\end{equation}
Note that each probability term is assumed to be independent, since they can be calculated irrelatively.

\textbf{I. Partition prior $p(\Pi_{[0,\tau]})$ \;\;\;} We assume
each frame is separately segmented and define the prior as,
\begin{eqnarray}
p(\Pi_{[0,\tau]})\!\!=\!\!\prod_{t=0}^\tau
p(\pi_t)\!\!=\!\!\prod_{t=0}^\tau\prod_{i=0}^{K_t}p(U_{t,i}).
\end{eqnarray}
Instead of using the Potts model as a partition prior in previous work~\cite{LinGraphMatching}, we predict the target location and size according to the scene surface property and information of camera calibration.

According to the research of using geometric context~\cite{TrackingSTContext}, the object size in the image plane is correlated with the physical size (in the real world) according to the scene geometric information, i.e. the camera parameters and the ground plane. The scene geometry can be roughly estimated in an interactive manner in a surveillance system according to a recent work~\cite{SurveillanceSystem}. We can then employ the informative prior of target size in the image plane, if the tracked targets belong to a specific object category. In other words, the prior distribution of target size is conditional on the target location in the image. In this work, considering the requirement of real-time processing, it is not practical to integrate the target recognition in the trajectory analysis, and we thus make the assumption that the semantic label of targets is specified in a certain scene. In fact, this assumption is reasonable, e.g., the indoor surveillance systems usually aim at about people while the outdoor systems usually track vehicles.

\begin{figure}[!htbp]
\begin{center}
 \includegraphics[width=0.55\linewidth]{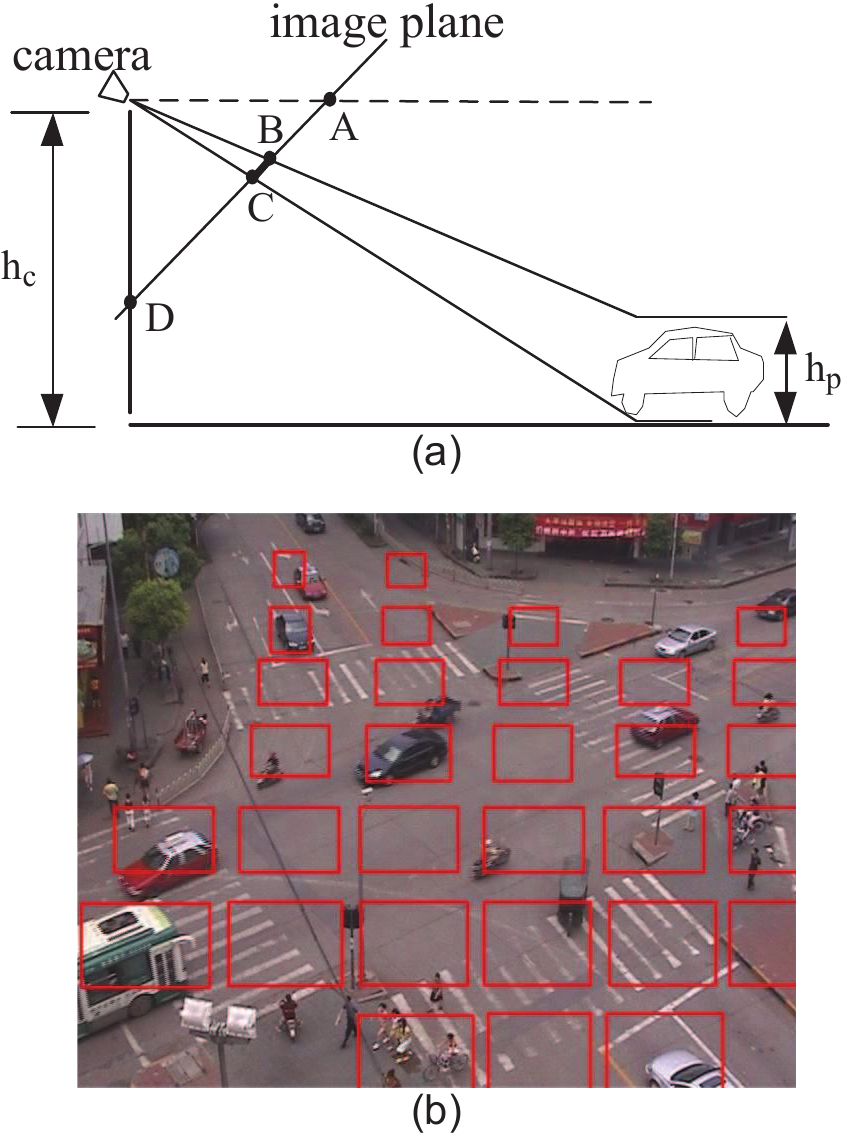}
\end{center}
 \caption{Location-size constraint.
(a) The target size in the surveillance image can be directly estimated according to the homography between the image plane and the ground plane; (b) We show an example of predicting vehicle sizes in the image as the prior information.
}\label{fig:locationSize}
\end{figure}

Fig.~\ref{fig:locationSize} (a) illustrates the location-size prediction with scene geometry. Let $B$ and $C$ denote the top and the bottom of car, $A$ the intersection of the car and the horizon line in the image plane, and $D$ the vertical vanishing point. Besides, let $h_p$ denote the
car height and $h_c$ the camera height. The expected size of
an observed vehicle on the ground plane can be predicted by simply following the cross ratio theorem,
\begin{equation}\label{equ:locationSize1}
\frac{BC}{BA} / \frac{DC}{DA} = \frac{h_p}{h_c-h_p}.
\end{equation}
Therefore we can obtain the target size distribution with respect to the target location $f_c(h,w|x,y)$. Suppose the location of target $U_{t,i}$ is $(x_{t,i}, y_{t,i})$ and the partition prior can be thus written as
\begin{eqnarray}
p(U_{t,i})  &\propto& f_c(h,w | x = x_{t,i}, y = y_{t,i}).
\end{eqnarray}
An example of predicting sizes of vehicles in the surveillance scene is presented in Fig.~\ref{fig:locationSize} (b), where we sample vehicle sizes from $f_c(h,w|x,y)$.

\begin{figure}[!htbp]
\begin{center}
\includegraphics[width=0.8\linewidth]{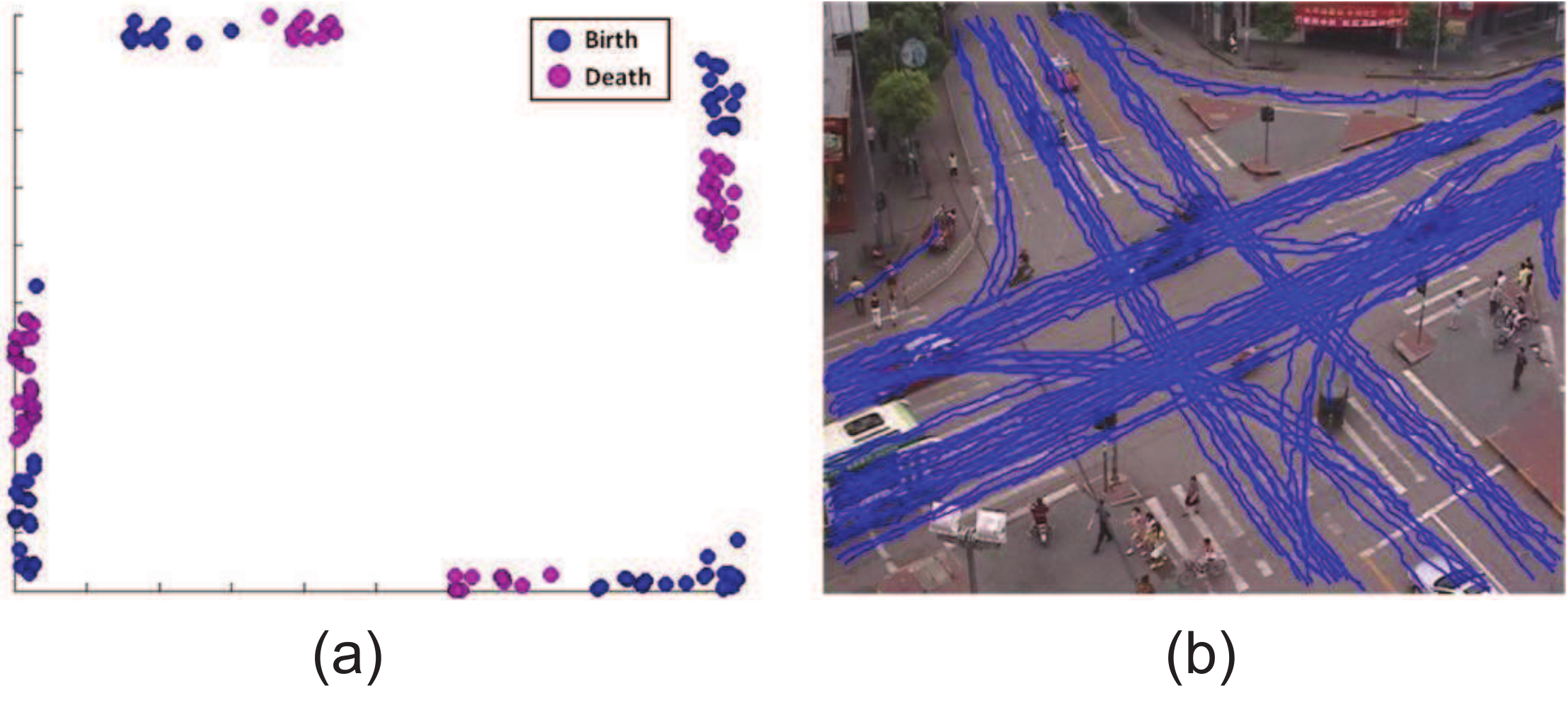}
\end{center}
\caption{A statistical path model for defining the matching prior on trajectory. (a) shows the statistical birth and death positions of moving targets in the scene; (b) shows the reference trajectories in the scene.
}\label{fig:pathmodel}
\end{figure}

\textbf{II. Matching prior on trajectory $p(\Phi_{[0, \tau]} )$ \;\;\;} For simplicity, we use the cable representation to define this prior model, which includes two terms: (i) the birth, death,
length (lifespan) of the cable, and (ii) trajectory shape of the cable. Thus, we have the matching prior factorized to obtain the following probability terms,
\begin{eqnarray}
p(\Phi_{[0, \tau]})\!&=&\!\prod_{i=0}^N p(C_i),\\
\!p(C_i )\! &=& \! p(t_{i,b},t_{i,d} )
p(\Gamma_i, \mathcal{R}),
\end{eqnarray}
where $C_i$ represents the $i$-th target trajectory. The first term $ p(t_{i,b},t_{i,d} )$ gives the prior distribution of birth/death on the global trajectory as shown in Fig.~\ref{fig:pathmodel} (a). $\Gamma_i$ denotes the trajectory shape, i.e. the curve of the trajectory. The second term $p(\Gamma_i, \mathcal{R})$ is a global motion prior based on a path model
$\mathcal{R}$, which consists of a set of reference trajectories $\{ \Gamma \}$, as shown in Fig.\ref{fig:pathmodel} (b). We can learn these reference trajectories by clustering in a supervised way according to the method reported by Wang et al~\cite{WangTrajectory}. Then the motion prior is in the form of a mixture model plus a robust statistic, as
\begin{eqnarray}
p(\Gamma_i, \mathcal{R}) \propto \exp \{ { - \min_{\Gamma_j \in \mathcal{R}} \Delta (\Gamma_i,\Gamma_j) + \epsilon } \},
\end{eqnarray}
where the function $\Delta(\cdot)$ denotes the geometrical distance~\cite{Procrustes} between the shapes of two trajectories, and $\epsilon$ is a tuning parameter for robustness.

\subsection{Likelihood Model}\label{subsect:likelihood}

The likelihood model $p(I_{[0,\tau]}|W_{[0,\tau]};\beta)$ includes the two following aspects: (i) the region appearances fitting with the background model, and (ii) the appearance consistency of the trajectories.
\begin{eqnarray}
\!\!\!\!\!\!\!p(I_{[0,\tau]}|W_{[0,\tau]};\beta)\!\!\!\! &=& \!\!\!\! \prod_{t=0}^\tau
 p(\Lambda^F_t | \pi_t, \mathcal{B})
 \\\nonumber
&&\cdot\prod_{i=1}^N p(\Lambda(C_i) | C_i),
\end{eqnarray}
where $\Lambda^F_t$ denotes the image domain of the foreground and $\mathcal{B}$  the background model proposed by ~\cite{BackgroundModel}. $\Lambda(C_i)$ indicates the image domain covered by trajectory $C_i$, i.e. the moving target $U_i$ over $t$ frames. The appearance consistency of the trajectories $p(\Lambda(C_i) | C_i)$ is equivalent to the matching similarity between targets over frames, as 
\begin{equation}
p(\Lambda(C_i)| C_i) = \prod_{t = t_{i,b}}^{t_{i,d}-1}
p(\Lambda^F_{t+1,i} | \Lambda^F_{t,i}),
\end{equation}
where $\Lambda^F_{t+1,i}$ and $\Lambda^F_{t,i}$ denote the image domains of adjacent targets. The target matching can be further calculated by measuring the composite features of the targets, 
\begin{equation}
p(\Lambda^F_{t+1,i} | \Lambda^F_{t,i}) \propto \exp \frac { {- \sum_{Z_i \in U_{t,i},\; Z_j \in U_{t+1, i}} E( Z_i, Z_j )}} {| U_{t, i} |},
\end{equation}
where $ E( v_i, v_j )$ is the distance metrics between two composite features, as defined in Eqn.\ref{eqn:feature_measure}. $| U_{t, i} |$ denotes the total number of extracted features in the target.


\section{Inference Algorithm}
\label{sect:inference}

Given the spatial and temporal graph representations, the problem of trajectory recovery is posed as two coupled tasks of spatial graph partitioning  $\Pi_{[0,\tau]}$ and temporal graph matching $\Phi_{[0,\tau]}$. In this section, we discuss a stochastic sampling algorithm to jointly solve the two tasks.


The reasons of using stochastic scheme rather than other deterministic optimization methods, e.g. Belief Propagation, or Graph-cuts, are as follow. (1) It is difficult to design fast searching rules due to the unpredictable variance and ambiguity of tracked targets. (2) The probabilistic formulation is a non-convex representation. (3) We usually cannot obtain the reliable initialization for trajectory analysis.

The proposed stochastic inference algorithm, designed under the Metropolis-Hasting mechanism~\cite{Metropolis}, is able to efficiently seek the optimal solution $W_{[0, \tau]}$ from the posterior probability $p(W_{[0, \tau]} | \I_{[0, \tau]})$ as defined in Eqn.~\ref{eq:posterior},
\begin{equation}
W_{[0, \tau]}^* \sim p(W_{[0, \tau]} | \I_{[0, \tau]}).
\end{equation}
We simulate a ergodic and aperiodic Markov Chain in which the algorithm visits a sequence of states in the joint space of $\{\Pi_{[0,\tau]}, \Phi_{[0,\tau]}\}$ over the time span $\tau$. Specifically, the sampling process iterates between two types of Markov Chain Monte Carlo (MCMC) dynamics and infers the graph partitioning $\Pi_{[0,\tau]}$ and graph matching $\Phi_{[0,\tau]}$ respectively. There are two components working in the iterative manner as follows:
\begin{itemize}
    \item Fixing the current state of graph matching $\Phi_{[0,\tau]}$, we  perform cluster sampling to explore the new solutions of graph partition $\Pi_{[0,\tau]}$. 
    \item Fixing the current state of graph partition $\Pi_{[0,\tau]}$, we update the graph matching state $\Phi_{[0,\tau]}$ by changing the matching relations of objects in the trajectories.
\end{itemize}
In both two components, each sampling is achieved by realizing a reversible jump (i.e. operator) between any two successive states to explore new solutions, for either graph partitioning or graph matching. The acceptance of a new state is decided based on a Metropolis-Hastings~\cite{Metropolis} decision to guarantee the convergence of the inference algorithm. In general, given two successive states $A$ and $B$ for either partitioning or matching, the acceptance rate is defined as:
\begin{equation}\label{eq:acceptance}
 \alpha( A \to B)  = \min\left(1, \frac{Q(B \to A) p( B )}{Q(A \to B)
 p(A)}\right),
\end{equation}
where $p(A)$ and $p(B)$ are the posterior probability of $W_{[0, \tau]}$ defined in Eqn.~\ref{eq:posterior}. $Q(B \to A)$ is the proposal probability to drive the state transition from $B$ to $A$ and conversely, $Q(A \to B)$ is the proposal probability from state $A$ to $B$.

How to design the proposal probability for driving the solution state transition is a non-trivial task that was addressed by a branch of works in  literature~\cite{SWCBarbu,MCMCDataAssociation,LinGraphMatching}. Recently, a MCMC-based cluster sampling algorithm, namely ``Swendsen-Wang Cut''(SWC), is proposed for image segmentation , which is able to simplify the  calculation of the ratio of proposal probability  $\frac{Q(B \to A)}{Q(A \to B)}$ in graphical models.  We refer to ~\cite{SWCBarbu} for the theoretical background.

In the following, we will discuss, respectively, the cluster sampling algorithm for graph partitioning and graph matching.



\begin{figure}[!htbp]
\begin{center}
\includegraphics[width=1.0\linewidth]{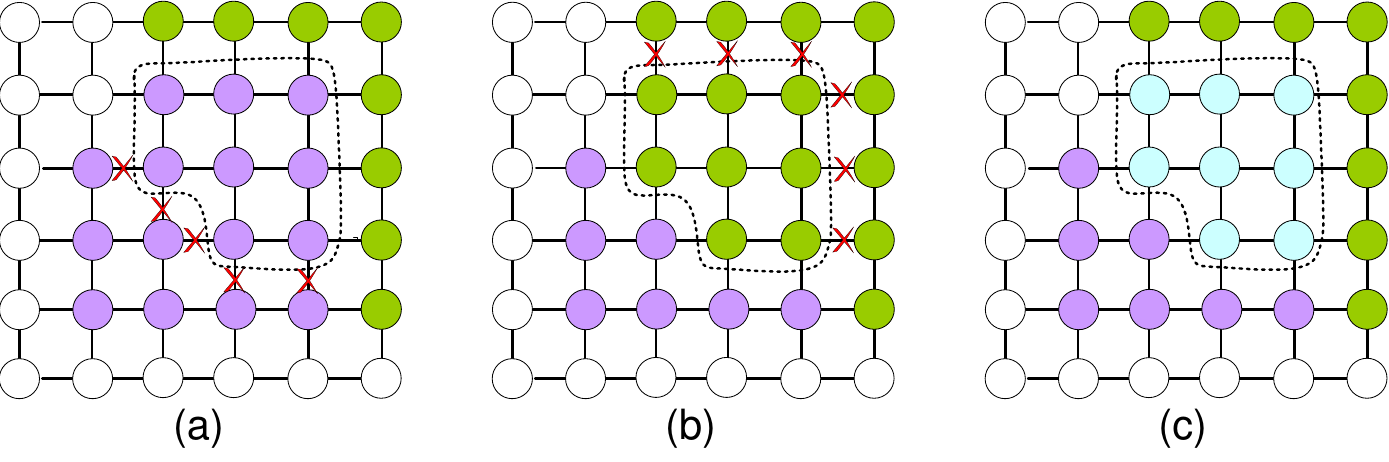}
\end{center}
\caption{Three typical solution states in spatial graph. At each stage of sampling for spatial graph partitioning, a connected cluster, $CC$, is generated  by turning edges off and then to be re-labeled for new solution states.
}\label{fig:spatial_states}
\end{figure}


\begin{figure}[!htbp]
\begin{center}
\includegraphics[width=1.0\linewidth]{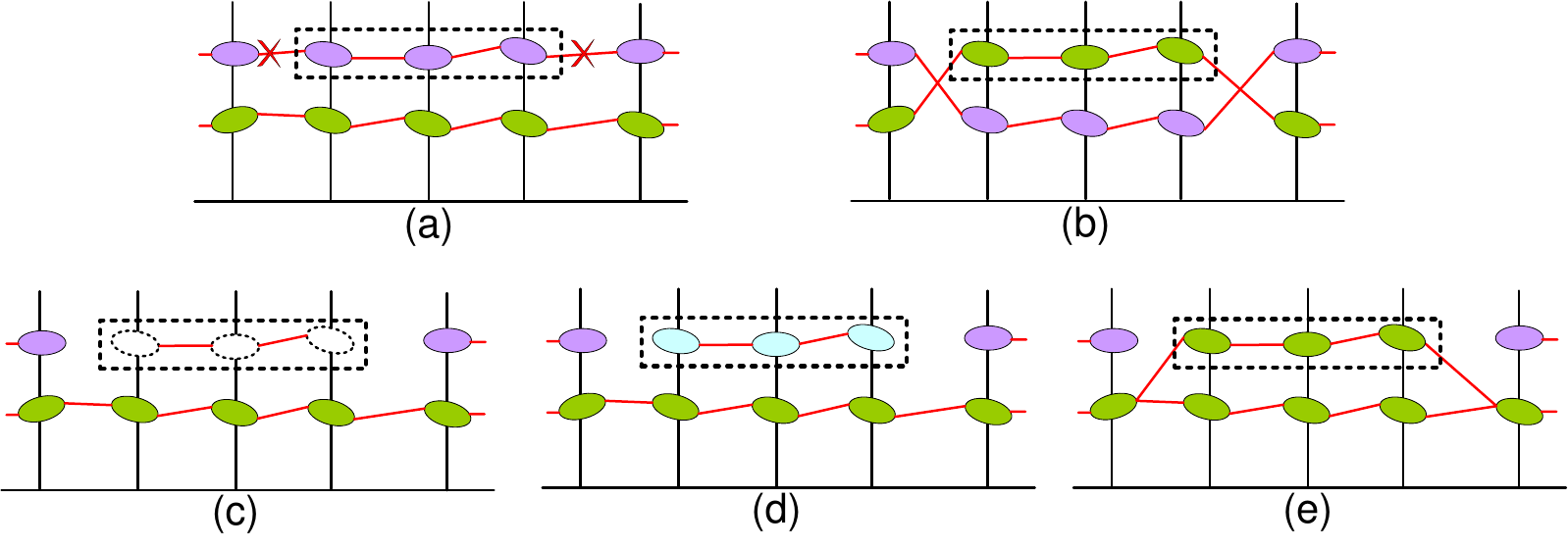}
\end{center}
\caption{Illustration of the inference in the temporal graph. (a) The connected cluster is generated by probabilistically turning off the edge connection. (b)-(e) show the solution state transition by different reversible jumps.
}\label{fig:temporal_states}
\end{figure}


\subsection{Sampling for Spatial Graph Partitioning}\label{sect:spatial_sampling}

Given a spatial graph $G^S_t$ extracted in the observed frame $\I_t, t\in [0, \tau]$, we utilize the SWC sampling for the graph partition inference. The algorithm achieves a reversible jump between two states in the solution space including the following two steps.

{\bf Step 1.} We generate a connected cluster by probabilistically turning off the edge links in the graph.

In the spatial graph $G_t^S = (V_t^S, E_t^S)$, suppose that $V_t^S$ is the set of graph vertices specifying the composite features and $E_t^S$ is the set of edges connecting neighboring graph vertices, as shown in Fig.~\ref{fig:stgraph} (c).   For notation simplicity, we omit the time stamp $t$ and the superscript $S$ in the algorithm description. For any edge $e \in E^S$, we introduce an auxiliary random variable $\mu_e = \{ \text{on} | \text{off} \}$, i.e. the connecting variable, which indicates whether the edge is turned on or off. The edge turn-on probability $q_e$ is defined according to the similarity of the two connected vertices,
\begin{equation}
q_e = p (\mu_e = on | v_a, v_b),
\end{equation}
where $v_a$ and $v_b$ are two graph vertices connected by the edge $e$. We collect some discriminative appearance and motion features (like the color, orientation gradient, and optical flow), which form a compact histogram $F$, i.e. each histogram bin indicates a specific feature dimension. For the image domain of the vertex, we describe colors by Luv metrics and pool over into $32$ bins; the orientation gradients are quantized with $48$ bins, and the optical flows with $9$ bins. For an edge $e = <v_a, v_b>$, the turn-on probability $q_e$ of two adjacent vertices can be thus defined with their appearance and motion consistency, as
\begin{eqnarray}\label{eq:qe}
     q_e \!\!\!&=&\!\!\! q(u_e= on |F(v_a),F(v_b)) \propto\\\nonumber
     && \!\!\!\!\!\!\!\!\exp \left(- \frac {\mathbf{K}(F(v_a) \| F(v_b)) + \mathbf{K}(F(v_b) \| F(v_a))} {T_e}\right),
\end{eqnarray}
where $\mathbf{K}( \cdot )$ is the Kullback-Leibler divergence between any two
histograms and $T_e$ is a constant temperature factor. 
Hence each edge is turned off with probability $1-q_e$ (as
shown in Fig.~\ref{fig:spatial_states}). It is worth mentioning that the turn-on probabilities of edges are calculated during the graph extraction before the sampling iteration.

For an arbitrary edge $e$, we then sample the connecting variable $\mu_e$ following the Bernoulli probability,
\begin{equation}
\mu_e \sim Bernoulli(q_e).
\end{equation}
Thus, graph vertices connected together by ``on'' edges form a connected cluster (denoted by $CC$ for simplicity), in which all vertices will share the same label in partitioning. Usually vertices in a $CC$ have similar appearance and thus most likely belong to
the same object. Fig.~\ref{fig:spatial_states} illustrates a $CC$ generated from different partition states. Note the edge between different objects (different colored nodes) are turned off deterministically. Compared to other graph partition algorithm (e.g., Graph-cuts~\cite{Graphcuts}) that turns off the edges by analytically finding the maximum flow over edges, the sampling method enables us to search for more possible solutions of graph partition.

Therefore, the ratio of proposal probability  $\frac{Q(B \to A)}{Q(A \to B)}$ in Eqn.~\ref{eq:acceptance} can be re-factorized as generating and labeling the connected cluster, as,
\begin{eqnarray}
\frac {Q(B \to A)} {Q(A \to B)} \!\!&=&\!\! \frac{q(CC^{B} | B) q(L(CC^{B})) }{q(CC^{A} | A) q (L(CC^{A}))},\\
\frac{q(CC^{B} | B)}{q(CC^{A} | A)} \!\!&=&\!\! \frac{ \sum_{e \in {\rm \mathbb{C}_B}} (1 - q_e)}{\sum_{e \in {\rm \mathbb{C}_A}}
(1 - q_e)},
\end{eqnarray}
where $CC^{A}$ and $CC^{B}$ denote the connected cluster generated on state $A$ and $B$, respectively. $\mathbb{C}_A$ denotes the set of edges that are turned off on state $A$, and similarly $\mathbb{C}_B$ is the turned-off edge set on $B$. Then we discuss the labeling of connected component in the next step.

{\bf Step 2.} We explore for a new solution of graph partitioning by labeling the generated $CC$. In practice, a few (e.g. $2 \sim 5$) $CC$s will be generated and we select one of them randomly.

Assume that the current partition state is $\Pi = \{ U_0, U_2, \ldots, U_K \}$ where $U_0$ denotes the background regions and $U_i, i \in [1, K]$ a segmented object. Note that the $CC$ may include the vertices from multiple targets. Then we can assign the $CC$ a label from $0$ to $K$ to update the partition state by three types of reversible jumps.
\begin{itemize}
    \item \textbf{Split-and-merge \;\;} The $CC$ is extracted from one object and merged into another one. The jump between the state (a) and (b) is an example as shown in Fig.~\ref{fig:spatial_states}. This jump is self-reversible.
    \item \textbf{Split \;\;} The selected $CC$ is assigned a new label, that is, a new object is created. In Fig.~\ref{fig:spatial_states}, from state (a) or (b)
to state (c) is a ``birth'' jump.
    \item \textbf{Merge \;\;} The whole object is selected as a $CC$ and merged into
another object, as from state (c) to state (a) or (b) in Fig.~\ref{fig:spatial_states}. The split jump and merge jump are mutual reversible.
\end{itemize}
These jumps can be defined in the same form as,
\begin{eqnarray}
\{ L(v) = i, v \in CC, i \in [1, K]\} \nonumber\\ \rightleftarrows \{ L(v) = i', v \in CC, i' \in [1, K]\},
\end{eqnarray}
where $L(v_j)$ indicates the label of vertex $v$.

\begin{algorithm}[ht!]\label{alg:sketch}
\caption{The sketch of trajectory analysis} 
\KwIn{A period of observed frames $[0,\tau]$, and $[\tau-3, \tau]$ frames are newly input.} 
\KwOut{The trajectory analysis solution $ W_{[0, \tau]} $.}

1. Construct graphs on new frames.\\
    ~~~~(1) Calculate an initial foreground map by the background subtraction.\\
    ~~~~(2) Extract the composite features by SURF and MSER detectors.\\
    ~~~~(3) Construct the initial spatial graphs on each frame with each composite feature being a vertex.\\
    ~~~~(4) Construct the initial temporal graph.\\
2. Perform the sampling algorithm with the new frames $[\tau-3, \tau]$.\\
    ~~~~(1) For each frame $t \in [\tau-3, \tau]$, loop for $80$ sampling iterations.\\
        ~~~~~~~~(i) Perform sampling for spatial graph partitioning on frame $t$.\\
        ~~~~~~~~(ii) Accept the new partition state according to the acceptance rate in Eqn.~\ref{eq:acceptance}.\\
    ~~~~(2) Sample the temporal graph matching with frames $[\tau-4, \tau]$ in $100$ iterations.\\
        ~~~~~~~~(i) Perform sampling for temporal graph matching.\\
        ~~~~~~~~(ii) Accept the new matching state according to the acceptance rate in Eqn.~\ref{eq:acceptance}.\\
3. Perform the sampling algorithm within the global observed period $[0, \tau]$.\\
    ~~Loop for $100$ Rounds\\
    ~~~~(1) Randomly select $3 \sim 5$ frames in $[0, \tau]$, and for each frame $t$ loop for $40$ sampling iterations. \\
    ~~~~~~~~(i) Perform sampling for spatial graph partitioning on frame $t$.\\
    ~~~~~~~~(ii) Accept the new partition state according to the acceptance rate in Eqn.~\ref{eq:acceptance}.\\
    ~~~~(2) Sample the temporal graph matching with frames $[0, \tau]$ in $100$ iterations.\\
        ~~~~~~~~(i) Perform sampling for temporal graph matching.\\
        ~~~~~~~~(ii) Accept the new matching state according to the acceptance rate in Eqn.~\ref{eq:acceptance}.\\
4. Output the final solution of trajectory analysis $W_{[0, \tau]}$.

\end{algorithm}

\subsection{Sampling for Temporal Graph Matching}\label{sect:temporal_sampling}

Graph matching sampling in the temporal graph is similar with sampling in the spatial graph. Note that the temporal sampling may cause state changing in the spatial graph, since each segmented object in the spatial graph is a node in the temporal graph, as shown in Fig.~\ref{fig:stgraph} (d).

Similarly, we first need to construct the temporal graph $G^T=(V^T,E^T)$ within the observed period $[0, \tau]$, and calculate the turn-on probabilities of edges $e^T \in E^T$ between arbitrary neighboring vertices. Recall that each vertex $v^T \in V^T$ indicates a moving target represented by a bounding box as shown in Eqn.~\ref{eq:target_represent}. We can thus use some simple appearance features on the image domains of vertices to define the turn-on probability, just similar with the definition in the spatial graph shown in Eqn.~\ref{eq:qe}. 

In the inference for graph matching, we first randomly select one trajectory $C_i$ at the current solution state, which is a bit different compared with the inference in the spatial graph. And we generate a sub-trajectory as the connected cluster $CC$ by probabilistically turning off the edge connections, as illustrated in Fig.~\ref{fig:temporal_states} (a). The $4$ types of reversible jumps are then performed to update the solution state. Fig.~\ref{fig:temporal_states} illustrates the transition of solution states.
\begin{itemize}
    \item \textbf{Birth \;\;} Assigning a new color for the selected $CC$, that is, to create a new cable (trajectory), as illustrated in Fig.~\ref{fig:temporal_states} (d). 

    \item \textbf{Merge \;\;} The selected $CC$ is merged into another cable, as shown in Fig.~\ref{fig:temporal_states} (e). In practice, we merge the $CC$ with neighboring cables. 

    \item \textbf{Death \;\;} Setting the selected $CC$ as background (false alarm), as shown in Fig.~\ref{fig:temporal_states} (c). 

\item \textbf{Swap \;\;} This is an important operator in temporal sampling. Given a selected $CC$, we swap it with another sub-cable in the same time span. Fig. ~\ref{fig:temporal_states} (b) is the succedent state of the current state in Fig. ~\ref{fig:temporal_states} (a) caused by the this operator. 

\end{itemize}
Assume that $N$ trajectories are traced in the observed period on the current state and each vertex $v$ in the trajectory represents a moving target. The birth, death, and swap jumps can be defined in the same form as,
\begin{eqnarray}
\{ L(v) = i , v \in CC, i \in [0, N]\} \nonumber \\ \rightleftarrows \{ L(v) = i', v \in CC, i' \in [0, N]\},
\end{eqnarray}
where $L(v)$ represents the label of $v$. The implementation for the swap jump is a bit different, since we need to select another sub-cable, as
\begin{eqnarray}
\{ L(v) \leftrightarrow  L(v'), v \in CC, v' \in CC' \} \nonumber \\ \rightleftarrows \{ L(v') \leftrightarrow  L(v), v' \in CC', v \in CC \},
\end{eqnarray}
where $L(v) \leftrightarrow  L(v')$ represents to swap labels of the two vertices.

We summarize the sketch of the proposed method in Algorithm~\ref{alg:sketch}, and introduce the detailed implementation in Section~\ref{sect:implementation}.

\subsection{Discussion of Convergence}

The joint space of $\{\Pi_{[0,\tau]}, \Phi_{[0,\tau]}\}$ over the time span $\tau$ is so large that it is prohibitive to search it exhaustively. For example, consider a case that there are $K$ spatial graph vertexes and $N$ trajectories (moving targets) in average. The solution space has in the order of $O((KN)^K)$. In statistics, we can simplify the maximum searching for joint probability by using the conditional probability, if the prior is assumed to be weak. This inspires us to design the algorithm to iteratively sample the conditional probabilities, $p( \Pi_{[0,\tau]} | \Phi_{[0,\tau]} )$ and $p(\Phi_{[0,\tau]} | \Pi_{[0,\tau]})$, respectively, with the two MCMC dynamics. The joint solution space is then separated into two relatively simple spaces.

For either solution space of spatial graph partitioning or temporal graph matching, the Markov chain is ergodic via performing the reversible jumps, based on the Metropolis-Hasting mechanism~\cite{Metropolis}. As the space is finite, all states can be visited following the observation that there is a non-zero probability for any node to be chosen into the connected component and assigned a label by activating the jumps. Then the Markov chain can move from a state to any other state with non-zero probability in finite steps.

In our method, we have to limit the number of sampling steps for efficiency consideration, as described in Algorithm~\ref{alg:sketch}. Then the global convergence is no longer guaranteed and the algorithm might obtain a local minimum. Nevertheless, we find the experimental results satisfactory due to the following reasons. First, the integration of informative prior models, e.g., $p(\Pi_{[0,\tau]})$, effectively accelerates the inference by fast rejecting false positive proposals. Second, the cluster sampling is much more efficient than traditional simpling methods. The process of generating the connected cluster is the key to efficiency improvement, in which the discriminative appearance and motion features are collected for generating effective proposals. Moreover, the cluster sampling enlarges the space that the stochastic process can possibly visit, and avoids often getting stuck in local minimums. An empirical study of inference convergence will be introduced in Section~\ref{sect:experiments}.

\section{Implementation}
\label{sect:implementation}

In this section, we apply our method to a video surveillance system which also involves a background modeling module~\cite{BackgroundModel}, and carry out the experiments with comparisons to the state-of-the-art approaches.

We start by introducing the parameter settings in our experiments. We set the value of the observed time span $\tau = 15$ frames, and we set the observed
window moving forward with a step-size of $\eta = 4$ frames. The other related parameters for our approach are introduced as follows. 

For the composite feature definition (in Section~\ref{sect:representation}), the histogram of local orientations $h(\cdot)$ consists of $72$ quantized bins and each bin indicates a small range of orientation angles, i.e. $5$ degrees. The weighted parameter $\lambda_g$ for measuring similarity of composite features is empirically set as $\lambda_g = 0.25$.

For the introduced prior models (in Section~\ref{sect:formuation}), we train them in an initial stage for each specific surveillance scene. The partition prior $p(\Pi_{[0, \tau]})$, i.e. the location-size prediction for tracked targets, is obtained by estimating the extrinsic camera parameters using an interactive calibration toolkit~\cite{SurveillanceSystem}, where we need to label a few parallel lines and tracked targets to calculate the vanishing points. Note that we make an assumption that the camera is fixed with only one degree of freedom, namely its height $h_c$. For the matching prior on trajectory $p(\Phi_{[0, \tau]} )$, we set the tuning parameter for robustness $\epsilon = 0.135$. The geometrical distance of two trajectories $\Delta$ is normalized into $[0,1]$. It is worth mentioning that we are allowed to disable these prior models by setting them uniform, although they are very effective in applications.

Given a period of observed frames $[0, \tau]$, we extract composite features on the newly arriving frames, i.e. $4$ frames for each sliding window, where we construct the spatial graphs and a temporal graph. Note that the initial temporal graph consists of composite features also, since temporarily no moving target is segmented in the new frames. In the following, the sampling procedure includes two stages: sampling in the new frames $[\tau-3, \tau]$ and sampling in the whole observed period $[0, \tau]$. 

(I) In the first stage, spatial graph partitioning is performed and the number of sampling iterations at each frame is bounded at $80$; vertices (composite features) are grouped to indicate potential moving targets due to their consistent appearances and motions. And then we sample the temporal graph matching with frames $[\tau-4, \tau]$, where the $(\tau-4)$-th frame should be taken into account, since we need to extract correspondences between the previous frames and the new frames. We set iteration number of the temporal matching sampling as $100$.

(II) In the second stage, the spatial graph partitioning and temporal graph matching are performed iteratively in a loop. The loop is set as $100$ rounds, and each round includes two sampling iterations. (1) First, a small number (i.e. $3 \sim 5 $) of frames in $[0, \tau]$ are first randomly selected for graph partition sampling, and the number of sampling iterations at each frame is bounded at $40$. (2) Then we perform matching sampling in the observed period $[0, \tau]$ for $100$ iterations.

\section{Experiments}
\label{sect:experiments}

We use three public video databases, TRECVID08~\cite{TRECVID}, PETS~\cite{PETS}, and LHI~\cite{LHIDataset}, to evaluate our method and compare with other state-of-the-arts approaches. These databases are very challenging for the multi-target tracking task, including scenarios with severe occlusions, scale changes or complex background structure. A number of video clips from these databases are selected for testing, i.e., $10$ videos from LHI, $8$ from PETS and $8$ from TRECVID. We manually annotate the bounding boxes of targets in the videos as the ground-truth. In our method, the types (semantic labels) of tracking targets are provided, which serve as the prior information. The videos selected from the TRECVID and PETS are all indoor scenes and the moving targets are all pedestrians; the videos in LHI are captured from outdoor traffic surveillance,  and we thus track the moving vehicles as the targets. Table~\ref{tab:databases} summarizes the number of frames as well as the number of targets in the testing videos.

All the testing videos are with the frame rate of $15$ fps and the frame size of $352 \times 288$ pixels. The experiments are carried out on a high-performance workstation with Core Duo $3.0$ GHZ CPU and $8$ GB memory. The computational efficiency for all steps (as described in Algorithm~\ref{alg:sketch}) in our system is summarized as follows. On average, the step of constructing graphs on new frames costs $80 \sim 100$ ms; it costs $300 \sim 450$ ms to perform sampling on new frames, including spatial graph partitioning and temporal graph matching; sampling within the global observed period costs around $600 \sim 800$ms. Recall that the algorithm processes $4$ newly arriving frames at a time, i.e., the observed window is moving with a step-size of $4$ frames. Thus, our system is capable of processing $3 \sim 5$ frames per second on average. In practice, we can enhance the efficiency by reducing the numbers of sampling iterations.

\begin{table}[!ht]
\centering
\begin{tabular}{|c|c|c|}
\hline
 Database   &  No. of Frames & No. of Targets \\
\hline
 TRECVID   &  $8972$  &  $389$ \\
\hline
 PETS &  $7409$  & $194$   \\
\hline
 LHI & $15213$ & $436$ \\
\hline
\end{tabular}
\caption{Testing sequences from public video databases}
\label{tab:databases}
\end{table}

\begin{figure*}[!ht]
\includegraphics[width=1.0\linewidth]{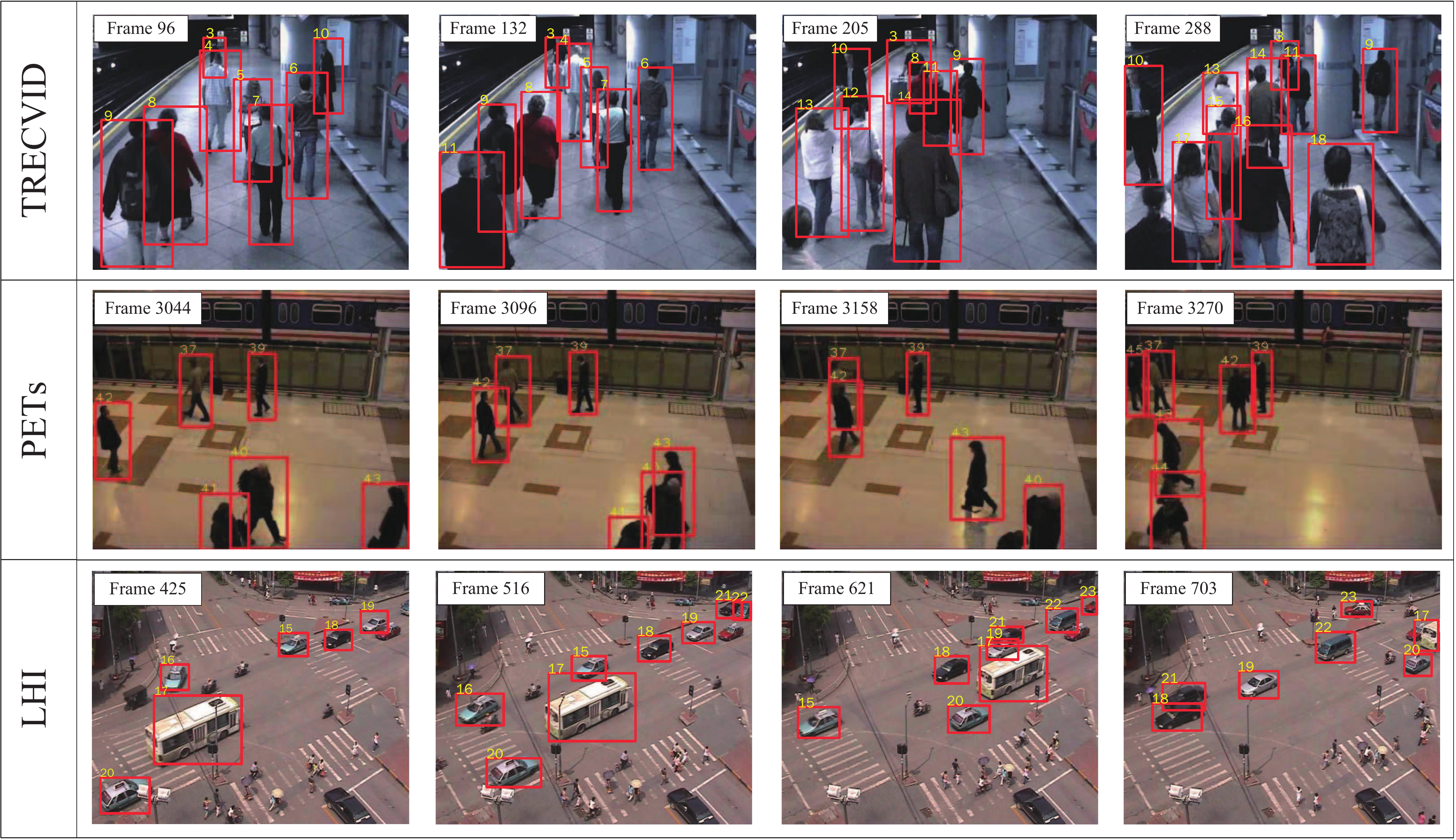}
\caption{Several representative tracking results on the public datasets. }\label{fig:tracking_result}
\end{figure*}

A few representative results of trajectory analysis are proposed in Fig.~\ref{fig:tracking_result}. Most of the video clips are very challenging due to the crowded objects, scale changes, severe occlusions and low resolution.

In order to quantitatively evaluate the performance, we introduce several object-level benchmark metrics, including {\bf Recall}, {\bf Precision}, {\bf FA/Frm}, and {\bf SwitchIDS}, as shown in Table~\ref{tab:metrics}, which are also adopted in~\cite{LearnAssociate2009,AdaptiveTracking}.  In the literature, some other performance measures have been proposed such as Multiple Object Tracking Precision and Accuracy (MOTA)~\cite{TrackHieHung,MCMCHuman}. These measures are less evident as they try to integrate multiple factors into one scalar valued measure, despite giving an overall picture of the performance. We write a program to match the results with the ground-truth based on these metrics automatically.

\begin{table*}[t]
\centering
\begin{tabular}{|c|c|}
\hline
{\bf Metric} & {\bf Definition}\\
\hline
Recall & Frame-based correctly matched targets / total ground-truth targets\\
\hline
Precision & Frame-based correctly matched targets / total output targets\\
\hline
FA/Frm & Frame-based number of false alarms per frame\\
\hline
SwitchIDS & The Number of times that the track IDs of two targets switch\\
\hline
\end{tabular}
\caption{Evaluation Metrics}
\label{tab:metrics}
\end{table*}

We compare our method with the recently proposed approaches for similar scenarios~\cite{TrackHieHung,TrackQBP,MCMCHuman}.

\begin{table}[!ht]
\centering
\begin{tabular}{|c|c|c|c|c|}
\hline
{\bf Methods} & {\bf Recall} & {\bf Precision} & {\bf FA/Frm} & {\bf SwitchIDS} \\
\hline
Zhao et al.~\cite{MCMCHuman} & $76.2\%$  & $ 72.7\% $ & $ 1.31 $ & $12$ \\
\hline
Huang et al.~\cite{TrackHieHung} & $ 69.1\% $  & $ 63.1\% $ & $ 1.82 $ & $13$\\
\hline
Leibe et al.~\cite{TrackQBP} & $ 78.9\% $ & $ 69.4 \%$ & $ 2.01$ & $9$\\
\hline
{\bf The proposed} & $83.3\%$ & $ 79.4\% $ & $ 0.72 $ & $7$ \\
\hline
{\bf without priors}  & $81.3\%$ & $ 78.2\% $ & $ 1.10 $ & $8$ \\
\hline
\end{tabular}
\caption{Results on videos from the TRECVID database}
\label{tab:result_tracvid}
\end{table}

\begin{table}[!ht]
\centering
\begin{tabular}{|c|c|c|c|c|}
\hline
{\bf Methods} & {\bf Recall} & {\bf Precision} & {\bf FA/Frm} & {\bf SwitchIDS} \\
\hline
Zhao et al.~\cite{MCMCHuman} & $82.4\%$  & $ 79.7\% $ & $ 0.92 $ & $18$ \\
\hline
Huang et al.~\cite{TrackHieHung} & $ 71.1\% $  & $ 68.5\% $ & $ 1.98 $ & $14$\\
\hline
Leibe et al.~\cite{TrackQBP} & $ 79.1\% $ & $ 73.1 \%$ & $ 1.38$ & $16$\\
\hline
{\bf The proposed} & $87.7\%$ & $ 82.9\% $ & $ 0.82 $ & $8$ \\
\hline 
{\bf without priors}  & $86.2\%$ & $ 79.8\% $ & $ 1.21 $ & $9$ \\
\hline
\end{tabular}
\caption{Results on videos from the PETS database}
\label{tab:result_PETS}
\end{table}

\begin{table}[!ht]
\centering
\begin{tabular}{|c|c|c|c|c|}
\hline
{\bf Methods} & {\bf Recall} & {\bf Precision} & {\bf FA/Frm} & {\bf SwitchIDS} \\
\hline
Huang et al.~\cite{TrackHieHung} & $ 73.2\% $  & $ 72.6\% $ & $ 1.27 $ & $14$\\
\hline
Leibe et al.~\cite{TrackQBP} & $ 79.7\% $ & $ 73.4 \%$ & $ 1.51$ & $10$\\
\hline
{\bf The proposed} & $91.3\%$ & $ 86.1\% $ & $ 0.84 $ & $7$ \\
\hline 
{\bf without priors}  & $90.8\%$ & $ 82.1\% $ & $ 1.07 $ & $9$ \\
\hline
\end{tabular}
\caption{Results on videos from the LHI database}
\label{tab:result_LHI}
\end{table}

\begin{figure*}[!ht]
\begin{center}
\includegraphics[width=0.75\linewidth]{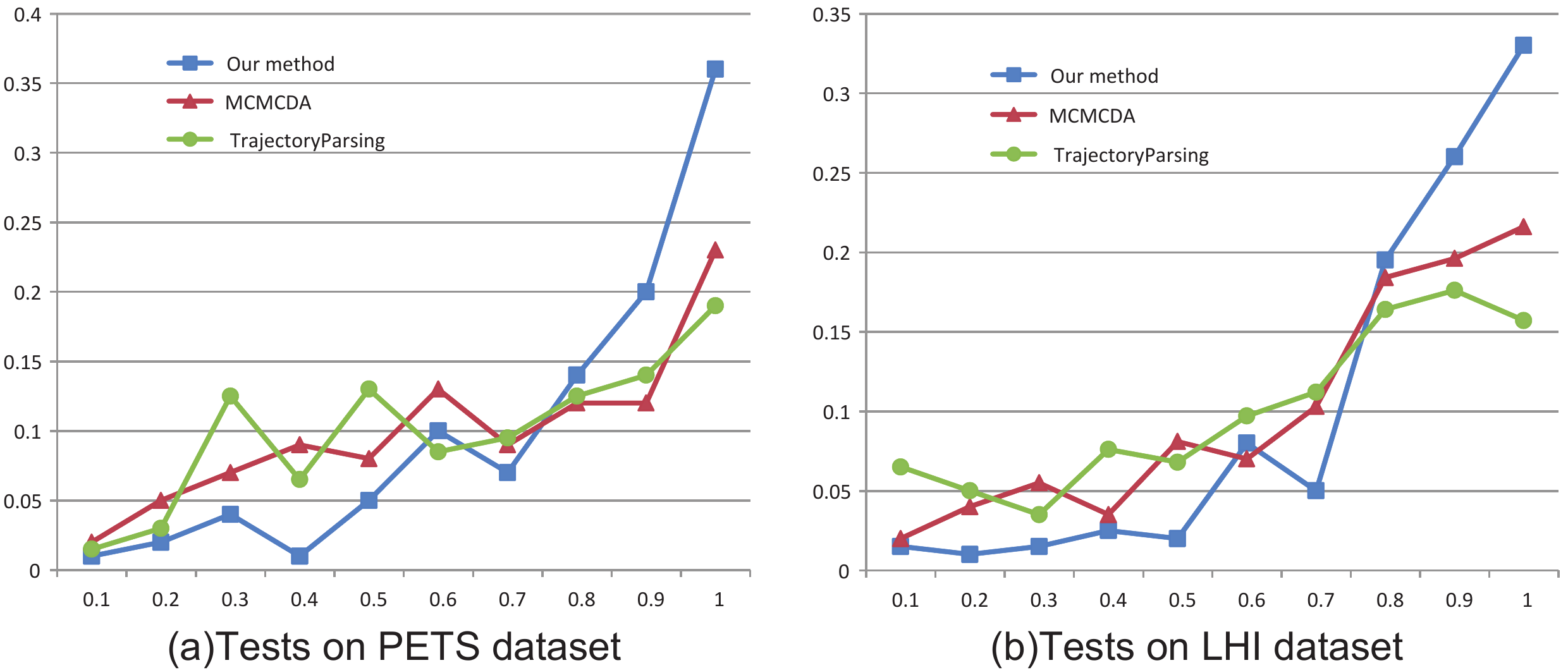}
\end{center}
\caption{The curves of Average Tracing Rate (ATR) for our trajectory analysis result and comparisons. The horizontal axis of ATR represents the coverage rate of the traced trajectory compared to the ground-truth; the vertical axis represents the proportion of trajectory length. In this evaluation, we compare our method with two other MCMC-based approaches: MCMC Data Association (MCMCDA)~\cite{MCMCDataAssociation} and Trajectory Parsing~\cite{TrajectoryParsing}. The curves on the left are tested on the PETS dataset, and the curves on the right are tested on the LHI dataset.}\label{fig:tracing_rate}
\end{figure*}

Table~\ref{tab:result_tracvid}, Table~\ref{tab:result_PETS}, and Table~\ref{tab:result_LHI} show the quantitative results of our results with the results proposed by Zhao et al.~\cite{MCMCHuman}, and Huang et al.~\cite{TrackHieHung}. The method by Zhao et al.~\cite{MCMCHuman} tracks pedestrians with a model-based approach to interpret the image observations by multiple partially occluded human hypotheses, and thus we only apply this method on the TRECVID and PETS databases for human tracking. The results show that our method achieves the best performance, greater Recall, greater Precision, fewer FA/Frm, and fewer SwitchIDs. To illustrate the benefits of using informative priors in trajectory analysis, we also report the system performances in the setting of disabling the prior components. The analysis of these experiments are presented as follows.

\begin{enumerate}
    \item Using deferred frames for global inference, i.e. an observed window, is very helpful, which provides us with more information to handle occlusions and mutual interactions.
    \item The prior components, e.g. the location-size prediction, gives very important cues for segmenting conglutinated targets; they effectively reduce the false alarms.
    \item The matching prior on trajectory, birth, death, lifespan of the cable, and shape of the cable, are strong constraints particularly for tracking vehicles in the traffic surveillance scene, since the motions of vehicles are usually regular in a certain scene.
    \item In the PETS dataset, many pedestrians have very similar appearances (e.g. in black coats) or motions (e.g. walking together), despite which the iterative sampling algorithm is shown to effectively reduce the number of SwitchIDs.
\end{enumerate}

In addition, we propose a novel benchmark metric to evaluate the trajectory-level performance, namely Average Tracing Rate (ATR), which is defined as the ratio of the traced trajectory length with respect to the ground-truth. The horizontal axis of ATR represents the coverage rate of the traced trajectory compared to the ground-truth of the testing videos; the vertical axis represents the proportion of trajectory length. The ATR for a result of trajectory analysis is in the form of a spot-curve for a discretized level of evaluation. This metric is very intuitive and straightforward to visualize the consistency of the tracking trajectories. In Fig.~\ref{fig:tracing_rate}, we propose the ATR curves of our method on the three datasets. In this evaluation, we compare with two other MCMC-based stochastic approaches for trajectory analysis, MCMC Data Association (MCMC) by Yu et al.~\cite{MCMCDataAssociation} and Trajectory Parsing by Liu et al. ~\cite{TrajectoryParsing}. 

\begin{figure*}[!ht]
\begin{center}
\includegraphics[width=0.9\linewidth]{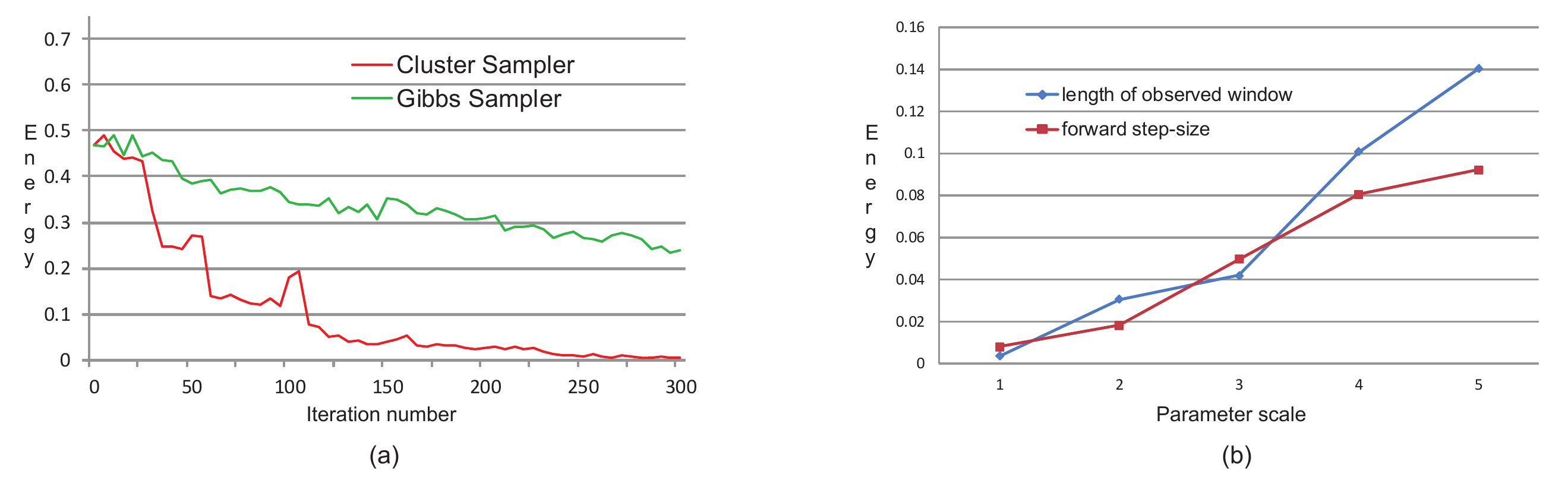}
\end{center}
\caption{Empirical study of algorithm convergence. (a) visualizes the energies (the vertical axis) of every $10$ iterations (the horizontal axis), i.e. $-\log(p(W_{[0,\tau]} | I_{[0,\tau]}))$, within an observed period $[0, \tau]$. The red curve and green curve in (a), respectively, represent the energies for our algorithm and the traditional Gibbs sampler. (b) shows the output converged energies with different parameters, the observed period length $\tau$ and the forward step-size $\eta$. The blue curve represents the converged energies with fixed $\eta = 4$ and increased $\tau$: $\tau = 15, 20, 25, 30, 35$. The red curve represents with fixed $\tau = 15$ and increased $\eta$: $\eta = 4, 6, 8, 10, 12$. }\label{fig:convergence}
\end{figure*}

To further analyze the algorithm convergence, we present an empirical study on visualizing the output energy in inference. Here the output energy, $-\log(p(W_{[0,\tau]} | I_{[0,\tau]}))$, is the logarithm of posterior probability within an observed period $[0, \tau]$. In Fig.~\ref{fig:convergence} (a), for an arbitrary period, we compare with Gibbs sampling for the trajectory analysis. For comparison, we replace the cluster sampling method at each step by the traditional Gibbs sampler~\cite{Gibbs} in the algorithm. We observe that the cluster sampling converges significantly faster. Moreover, we investigate the output energies with respect to the two important parameters in our system, the  observed period length $\tau$ and the forward step-size $\eta$. This experiment is also carried out within a period of observed frames. We first fix $\eta=4$ and discretely increase $\tau$ by $5$ scales: $\tau = 15, 20, 25, 30, 35$. That is, we increase the length of period and deal with more video frames in inference. Then we increase $\eta = 4, 6, 8, 10, 12$ with fixed $\tau=15$, to gradually reduce the overlap with the previous inference. The empirical results are reported in Fig.~\ref{fig:convergence} (b), where the horizontal axis represents the scale for either parameter.

\section{Conclusion}
\label{sect:discussion}

The objective of this paper is to track multiple video targets and recover their trajectories, against occlusion, interruption, and background
clutter. Compared with the previous methods in literature, the main contributions of this paper are as follows. First, we propose a novel unified framework of trajectory analysis to together solve spatial graph partitioning and temporal graph matching. Second, a robust composite feature bundling the MSER feature and SURF feature is presented for the affinity model of moving targets, against scale transition and non-rigid motion. Third, we design a stochastic sampling algorithm to iteratively solve the spatial graph partition and temporal graph matching. This algorithm is designed under the Metropolis-Hastings method without the need for good initializations.

We have applied our method in an intelligence video system and found satisfactory performance. In experiments, our method is tested on several challenging videos from the public video databases of visual surveillance, including TRECVID, PETS, and LHI, and it outperforms the state-of-the-art methods.

In future work, it is important to integrate object recognition~\cite{LinPR} into the trajectory analysis, which will lead to a more general solution for video surveillance applications. In addition, we plan to study the parallel implement for the MCMC-based inference to further improve the computation efficiency.



\begin{thebibliography}{00}


\bibitem{EnsembleTrack}
S. Avidan, Ensemble Tracking, {\em IEEE Trans. on PAMI}, 29(2): 261-271, 2007.

\bibitem{MILTrack}
B. Babenko, M. H. Yang, and S. Belongie, Visual Tracking with Online Multiple Instance Learning, {\em Proc. IEEE Conference on Computer Vision and Pattern Recognition}, 2009.

\bibitem{SWCBarbu}
A. Barbu and S. C. Zhu, Generalizing Swendsen-Wang for Image Analysis, {\em
Journal of Computational and Graphical Statistics}, 16(4): 877-900, 2007.

\bibitem{JPDAF}
Y. Bar-Shalom, T. Fortmann, and M. Scheffe, Joint probabilistic data association for multiple targets in clutter, {\em Information Sciences and Systems}, 1980.

\bibitem{SURF}
H.t Bay, A. Ess, T. Tuytelaars, and L. V. Gool, SURF: Speeded Up Robust Features, Computer Vision and Image Understanding (CVIU), 110(3): 346-359, 2008.

\bibitem{SMC-Robot}
N. Bellotto, and H. Hu, Multisensor-Based Human Detection and Tracking for Mobile Service Robots, {\em IEEE Transactions on SMC, Part B}, 39(1): 167-181, 2009.

\bibitem{TrackDP}
J. Berclaz, F. Fleuret, and P. Fua, Robust People Tracking with Global Trajectory Optimization, {\em Proc. IEEE Conference on Computer Vision and Pattern Recognition}, 2006.

\bibitem{Graphcuts}
Y. Boykov, O. Veksler, and R. Zabih, Fast Approximate Energy Minimization via Graph Cuts, {\em Proc. of IEEE International Conference on Computer Vision}, 1999.

\bibitem{SMC-PartTracking}
Y. Xie, L. Lin, and Y. Jia, Tracking Objects with Adaptive Feature Patches for PTZ Camera Visual Surveillance, {\em Proc. of IEEE International Conference on Pattern Recognition}, 2010.






\bibitem{OpticalFlow}
S. P Denman, V. Chandran, and S. Sridharan, An Adaptive Optical Flow Technique for Person Tracking Systems, {\em Pattern Recognition Letters}, 28(10): 1232-1239, 2007.

\bibitem{Procrustes}
I. Dryden and K. Mardia, Statistical Shape Analysis, {\em John Wiley and Sons}, 1998.

\bibitem{PETS}
R. B. Fisher, The PETS04 Surveillance Ground-truth Data Sets, {\em Proc. IEEE International Workshop on Performance Evaluation of Tracking and Surveillance }, 2004.

\bibitem{Gibbs}
S. Geman and D. Geman, Stochastic Relaxation, Gibbs Distributions, and the Bayesian Restoration of Images. {\em IEEE Transaction on Pattern Anal. and Mach. Intel.}, 6(6):721-741, 1984.



\bibitem{TrackHieHung}
C. Huang, B. Wu, and R. Nevatia, Robust Object Tracking by Hierarchical Association of Detection Responses, {\em Proc. European Conference on Computer Vision}, 2008.

\bibitem{TIP-AdaptiveTracking}
N. Jiang, W. Liu, and Y. Wu, Learning Adaptive Metric for Robust Visual Tracking, {\em IEEE Transactions on Image Processing}, 20(8): 2288-2300, 2011.

\bibitem{Hungarian}
R. Jonker and A. Volgenant, A Shortest Augmenting Path
Algorithm for Dense and Sparse Linear Assignment Problems. {\em
Computing}, 38, 325--340, 1987.



\bibitem{SMC-Kalman}
A. Kiruluta, M. Eizenman, S. Pasupathy, Predictive Head Movement Tracking Using a Kalman Filter, {\em IEEE Transactions on SMC, Part B}, 27(2): 326-331, 1997.


\bibitem{MCMCParticleFiltering}
Z. Khan, T. Balch, and F. Dellaert, An MCMC-based Particle Filter
for Tracking Multiple Interacting Targets, {\em Proc. of IEEE Conference on Computer Vision and Pattern Recognition}, 2009.

\bibitem{TrackQBP}
B. Leibe, K. Schindler, and L. V. Gool, Coupled Detection and Trajectory Estimation for Multi-Object Tracking, {\em Proc. Intel' Conference on Computer Vision}, 2007.

\bibitem{LearnAssociate2009}
Y. Li, C. Huang, and R. Nevatia, Learning to Associate: HybridBoosted Multi-Target Tracker for Crowded Scene, {\em Proc. Intel' Conference on Computer Vision}, 2009.

\bibitem{LinGraphMatching}
L. Lin, X. Liu and S. C. Zhu, Layered Graph Matching with Composite Cluster Sampling, {\em IEEE Trans. on PAMI}, 32(8): 1426-1442, 2010.

\bibitem{LinEvent}
L. Lin, H. Gong, L. Li, and L. Wang, Semantic Event Representation and Recognition Using Syntactic Attribute Graph Grammar, {\em Pattern Recognition Letters}, 30(2): 180-186, 2009.


\bibitem{LinPR}
L. Lin, P. Luo, X. Chen, and K. Zeng, Representing and Recognizing Objects with Massive Local Image Patches, {\em Pattern Recognition}, 45(1): 231-240, 2012.

\bibitem{LinTrackingAR}
L. Lin, Y. Wang, Y. Liu, C. Xiong and K. Zeng, Marker-less Registration Based on Template Tracking for Augmented Reality, {\em Multimedia Tools and Applications}, 41(2): 235-252, 2009.

\bibitem{SurveillanceSystem}
X. Liu, L. Lin, H. Jin, S. Yan, and W. Tao, Integrating Spatio-temporal Context with Multiview Representation for Object Recognition in Visual Surveillance, {\em IEEE Transactions on Circuits and Systems for Video Technology}, 21(4): 393-407, 2011. 


\bibitem{TrajectoryParsing}
X. Liu, L. Lin, S.C. Zhu, and H. Jin, Trajectory Parsing by Cluster Sampling in Spatio-Temporal Graph, {\em Proc. of IEEE Conference on Computer Vision and Pattern Recognition}, 2009.


\bibitem{AdaptiveTracking}
X. Liu, L. Lin, S. Yan, H. Jin, and W. Jiang, Adaptive Object Tracking by Learning Hybrid Template On-line, {\em IEEE Trans. on Circuits and Systems for Video Technology}, 21(11): 1588-1599, 2011.

\bibitem{TIP-Context}
E. Maggio, A. Cavallaro, Learning Scene Context for Multiple Object Tracking, {\em IEEE Transactions on Image Processing}, 18(8): 1873-1884, 2009.

\bibitem{MSER}
J. Matas, O. Chum, M. Urban, and T. Pajdla, Robust Wide
Baseline Stereo from Maximally Stable Extremal Regions, {\em Proc. British Machine Vision Conference}, 384--393, 2002.

\bibitem{Metropolis}
N. Metropolis, A.W. Rosenbluth, M.N Rosenbluth, A.H. Teller, E. Teller, Equation of state calculations by fast computing machines, {\em Journal of Chemical Physics}, 21(6): 85-111, 1953.

\bibitem{SMC-DecisionTrack}
T. K. Moon, S. E. Budge, W. C. Stirling, J. B. Thompson, Epistemic Decision Theory Applied to Multiple-Target Tracking, {\em IEEE Transactions on SMC, Part B}, 24(2): 234-245, 1994.

\bibitem{TrackingSTContext}
H. T. Nguyen, Q. Ji, and A. W. M. Smeulders, Robust Multi-Target Tracking Using Spatio-Temporal Context, {\em Proc. IEEE Conference on Computer Vision and Pattern Recognition}, 2006.

\bibitem{TIP-Surveillance}
O. Barnich, M. Van Droogenbroeck, ViBe: A Universal Background Subtraction Algorithm for Video Sequences, {\em IEEE Transactions on Image Processing}, 20(6): 1709-1724, 2011.

\bibitem{TIP-SeqParticle}
P. Pan, D. Schonfeld, Video Tracking Based on Sequential Particle Filtering on Graphs, {\em IEEE Transactions Image Processing}, 20(6): 1641-1651, 2011.

\bibitem{MHT}
D. Reid, An algorithm for Tracking Multiple Targets. {\em TAC}, 24(6): 84-90, 1979.

\bibitem{TIP-MultipleTrack}
J. Seong-Wook, R. Chellappa, A Multiple-Hypothesis Approach for Multiobject Visual Tracking, {\em IEEE Transactions on Image Processing}, 16(11): 2849-2854, 2007.

\bibitem{TRECVID}
A. Smeaton, P. Over, and W. Kraaij, Evaluation Compaings and Trecvid, {\em Proc. ACM international workshop on Multimedia information retrieval}, pp. 321-330, 2006.

\bibitem{BackgroundModel}
Y. Zhao, H. Gong, L. Lin, and Y. Jia, Spatio-temporal Patches for Night Background Modeling by Subspace Learning, {\em Proc. of IEEE International Conference on Pattern Recognition}, 2008.




\bibitem{TuytelaarsFeatureSurvey}
T. Tuytelaars and K. Mikolajczyk, Local Invariant Feature Detectors: A Survey, {\em Foundations and Trends in Computer Graphics and Vision}, 3(2): 177-280, 2007.




\bibitem{WangTrajectory}
X. Wang and K. Tieu and W. E. L. Grimson, Learning Semantic Scene Models
by Trajectory Analysis, {\em Proc. European Conference on Computer Vision}, Vol.3: 110-123, 2006.

\bibitem{BundledFeature}
Z. Wu, Q. Ke, M. Isard, and J. Sun, Bundling features for large scale partial-duplicate web image search, {\em Proc. IEEE Conference on Computer Vision and Pattern Recognition}, vol. 1: 25-32, 2009.


\bibitem{SMC-DynamicProg}
L. Yang, J. Si, K. S. Tsakalis, A. A. Rodriguez, Direct Heuristic Dynamic Programming for Nonlinear Tracking Control With Filtered Tracking Error, {\em IEEE Transactions on SMC, Part B}, 39(6): 1617-1622, 2009.

\bibitem{LHIDataset}
B. Yao, X. Yang, L. Lin, M. Lee, and S.C. Zhu, I2T: Image Parsing to Text Description, {\em Proceeding of IEEE}, 98(8): 1485-1508, 2010.

\bibitem{TrackingSurvey}
A. Yilmaz and O. Javed, Object Tracking: A Survey, {\em ACM Computing Survey}, 38(4), 13, 2006.

\bibitem{MCMCDataAssociation}
Q. Yu, G. Medioni, and I. Cohen, Multiple Target Tracking Using
Spatio-Temporal Markov Chain Monte Carlo Data Association, {\em Proc. of IEEE Conference on Computer Vision and Pattern Recognition}, 2007.



\bibitem{MCMCHuman}
T. Zhao, R. Nevatia, and B. Wu, Segmentation and Tracking of Multiple Humans in Crowded Environments, {\em IEEE Trans. on PAMI}, 30(7): 1198 - 1211, 2008.




 \end{thebibliography}
\end{document}